\documentclass[sigconf]{acmart}

\usepackage{algorithm}
\usepackage{algpseudocode}  
\usepackage{amsmath}  
\def\BibTeX{{\rm B\kern-.05em{\sc i\kern-.025em b}\kern-.08emT\kern-.1667em\lower.7ex\hbox{E}\kern-.125emX}}
    
\copyrightyear{2019}
\acmYear{2019}
\setcopyright{acmlicensed}
\acmConference[KDD '19]{KDD '19: The 25th ACM SIGKDD Conference on Knowledge Discovery \& Data Mining}{August 04--08, 2019}{Anchorage, Alaska USA}
\acmBooktitle{KDD '19: The 25th ACM SIGKDD Conference on Knowledge Discovery \& Data Mining, August 04--08, 2019, Anchorage, Alaska USA}
\acmPrice{15.00}
\acmDOI{10.1145/nnnnnnn.nnnnnnn}
\acmISBN{978-x-xxxx-xxx-x/YY/MM}

\begin{document}

%
\title{Deep Hierarchical Reinforcement Learning Based \\Recommendations via Multi-goals Abstraction}

%
\author{Dongyang Zhao}
\affiliation{%
 \institution{School of Electronics Engineering and Computer Science}
 \institution{Peking University}
 }
 \email{zdy_macs@pku.edu.cn}

\author{Liang Zhang}
\affiliation{
  \institution{JD.com}
}
\email{zhangliang16@jd.com}
\author{Bo Zhang}
\affiliation{
  \institution{JD.com}
}
\email{zhangbo35@jd.com}

\author{Lizhou Zheng}
\email{zhenglizhou1@jd.com}
\affiliation{
  \institution{JD.com}
}
\author{Yongjun Bao}
\affiliation{
  \institution{JD.com}
}
\email{baoyongjun@jd.com}
\author{Weipeng Yan}
\affiliation{
  \institution{JD.com}
}
\email{Paul.yan@jd.com}

\renewcommand{\shortauthors}{Dongyang Zhao and Liang Zhang, et al.}
%
\begin{abstract}
The recommender system is an important form of intelligent application, which assists users to alleviate from information redundancy. Among the metrics used to evaluate a recommender system, the metric of conversion has become more and more important. The majority of existing recommender systems perform poorly on the metric of conversion due to its extremely sparse feedback signal. To tackle this challenge, we propose a deep hierarchical reinforcement learning based recommendation framework, which consists of two components, i.e., high-level agent and low-level agent. The high-level agent catches long-term sparse conversion signals, and automatically sets abstract goals for low-level agent, while the low-level agent follows the abstract goals and interacts with real-time environment. To solve the inherent problem in hierarchical reinforcement learning, we propose a novel deep hierarchical reinforcement learning algorithm via multi-goals abstraction (HRL-MG).
Our proposed algorithm contains three characteristics: 1) the high-level agent generates multiple goals to guide the low-level agent in different stages, which reduces the difficulty of approaching high-level goals; 2) different goals share the same state encoder parameters, which increases the update frequency of the high-level agent and thus accelerates the convergence of our proposed algorithm; 3) an appreciate benefit assignment function is designed to allocate rewards in each goal so as to coordinate different goals in a consistent direction. We evaluate our proposed algorithm based on a real-world e-commerce dataset and validate its effectiveness.
\end{abstract}



\maketitle
\section{Introduction}
In this information era, end users/consumers usually suffer from heavy burden of content and product choices when browsing the Internet. The recommender system is an important form of intelligent application, which assists users to alleviate from such information redundancy and save time of picking up what they want from lots of irrelevant contents and products. More specifically, the recommender agents discover users' short-term and long-term interests/ preferences from their browsing histories in Internet, e.g., products, news, movies and music, as well as various types of services\cite{Paul1997,Francesco2011}. They build user models based on their interests/preferences and automatically recommend personalized items so as to satisfy users' information needs. As a result, the recommender systems have become increasingly popular, and have been applied to a variety of domains in Internet, e.g., e-commerce, news, movies, etc. 

To improve the performance of recommender systems, lots of works have been proposed, evolving from the traditional shadow models like the collaborative filtering model\cite{John1998}, to the mainstream deep models like the wide\&deep model\cite{widedeep} and finally to the trend of deep reinforcement learning based methods \cite{Zhao2018}. The deep neural networks have shown excellent performance, due to their powerful capabilities of extracting features and relationships. For instance, DIEN\cite{Guorui2018} designed a interest extractor layer to capture temporal interests from historical behavior sequence. Most of these deep methods are static, which can hardly follow the dynamic changes of users' preferences. The deep reinforcement learning (DRL) based methods overcome this problem via interacting with users in real time and dynamically adjust the recommendation strategies. For instance, DEERS \cite{Zhao2018} adopted a Deep Q-Network framework and integrated both positive and negative feedback simultaneously. Furthermore, the DRL based recommendations maximize the long-term cumulative expected returns, instead of just immediate (short-term) rewards as traditional deep model, which can bring more benefits in the future.

At present, the majority of works about recommender systems focus on optimizing the metric of click and have already achieved great improvements. As the competition becomes fiercer, the recommender agents gradually pay more attention on the metric of conversion, especially in e-commerce recommender systems. On the one hand, the metric of conversion is more realistic as counterfeiting conversion is more difficult. On the other hand, the e-commerce recommender systems usually recommend natural items and display ads together. The advertisers care more about the direct conversions, instead of indirect clicks, so as to guarantee their revenue over investment. Few of works consider the metric of conversion. For instance, Yang et al.\cite{Yang2016} combined natural language processing and dynamic transfer learning into a unified framework for conversion rate (CVR) prediction. Either of these works only optimize the metric of click or the metric of conversion. The click and conversion are highly correlated, but may not have the positive correlation. An item which is more likely to be clicked, may results in lower probability of conversion, e.g., the item with relative cheap price but poor product quality.

In this paper, we adopt deep reinforcement learning based methods to optimize the metrics of click and conversion jointly. The user behaviors can be treated as a sequential pattern, i.e., from impression, to click and finally conversion. More specifically, when a list of recommended items are exposed to users, users may click some items in which they are interested, and then buy the favorite items. This pattern reflects users' hierarchical interests. The click signals from part of exposed items reflect various superficial interests such as the curiosity for new items, the return clicks for some previously purchased items, the initial purchase willingness, etc, while the conversion signals from part of clicked items show the pure and deep purchase interests. As a result, the conversion signals are much sparser than the click signals. The existing deep reinforcement learning based methods in recommendations usually treat the conversion signals just as same as the clicks, except for assigning some large weights. For instance, Hu et al.\cite{Yujing2018} assigned the conversion weight according to the price of each product item. The large weights can partially alleviate the sparsity problem of conversion signals. Yet such method requires deep reinforcement learning techniques to track the conversion signals from impressions directly, just as tracking click signals from impressions. This makes sparse conversion signals more likely to be covered by click signals.

To solve this sparsity problem, we propose a deep hierarchical reinforcement learning based recommendation framework, which consists of two components, i.e., high-level agent and low-level agent. More specifically, the high-level agent tries to catch the long-term sparse conversion signals based on users' click and conversion histories. The actor of the high-level agent automatically sets goals for the low-level agent. On the other hand, the low-level agent captures the short-term click signals based on users' impression and click histories. The actor of low-level agent interacts with the real-time environment via making actual recommendations and receiving feedback from users. This framework differentiates the hierarchical interests in users' behavior patterns via hierarchical agents. There exist several problems in this hierarchical reinforcement learning framework. Firstly, how does the high-level agent automatically generate goals for the low-level agent. The high-level goals affect the performance of the framework significantly, but there exists no explicit goals for the high-level agent in recommender systems. Secondly, how does the high-level goals influence the low-level agent. The appropriate way to guide the low-level agent can reduce the difficulty of approaching the high-level goals. Thirdly, how to increase the update frequency of high-level agent so as to accelerate its convergence. The feedback frequency of high-level agent is far less than that of low-level agent. 

To tackle these challenges, we further propose a novel deep hierarchical reinforcement learning algorithm (HRL-MG), in which the high-level agent guides the low-level agent via multi-goals abstraction. In the interaction between recommender agents and users, the high-level agent first generates a set of abstract goals based on users' click and conversion histories, and conveys them to the low-level agent. Each abstract goal has the same form as the action of the low-level agent. Furthermore, different abstract goal guides the low-level agent in different interaction stage. All these make the high-level goals easier to follow and approach. Then, the low-level agent generates actual recommendation items based on users' browsing and click histories, and collects users' feedback as external reward. The low-level agent also accepts the internal reward, which is generated from the difference between the action and its corresponding goal. Finally, the low-level agent conveys the users' feedback to the high-level agent to improve the quality of different goals. To enhance the cooperation of each goal, we design the same state encoder structure for each goal, the parameters of which are also shared by all goals. These parameters are updated when each goal updates its own parameters. In addition, we design an appreciate reward mechanism based on users' feedback, called benefit assignment function, to coordinate the goals in a consistent direction.

In summary, this paper has the following contributions:
\begin{itemize}
\item To the best of our knowledge, we are the first to propose a DHRL based recommendation framework. The high-level agent catches the long-term sparse conversion signals, while the low-level agent captures the short-term click signals. \
\item We propose a novel deep hierarchical reinforcement learning algorithm (HRL-MG), in which the high-level agent guides the low-level agent via multi-goals abstraction. The multiple high-level goals reduce the difficulty for the low-level agent to approach the high-level goals. \
\item We design a shared state encoder for each goal so as to accelerate the update frequency and an appreciate benefit assignment function to allocate rewards in each goal so as to coordinate different goals correctly. 
\item We carry out the offline and online evaluation based on the real-world e-commerce dataset from JD.com. The experimental results demonstrate the effectiveness of our proposed algorithm.
\end{itemize}

In this paper, we first introduce the details of our proposed framework in Section 2. Then, we present our training procedure in Section 3. After that, we demonstrate our experiments in Section 4. The related work is discussed in Section 5. At last, we conclude this paper in Section 6.

\section{The proposed framework} 
This section begins with an overview of the proposed recommendation framework based on hierarchical reinforcement learning. 
Then we introduce the technical details of the high-level agent and the low-level agent.

\subsection{Framework Overview}
As mentioned above, we model the recommendation task as a Markov Decision Process(MDP) and leverage the techniques of reinforcement learning to automatically learn the optimal recommendation strategy. 
Users are regarded as the environment, and recommendation system is regarded as the agent.
Users' preferences are the environment state in which the agent is located. 
According to current state, the agent select an action (giving corresponding recommended item), and then the environment gives feedback: skip, click, order(convert), or leave, etc.
The recommendation agent obtains corresponding reward, and the state of the environment is updated, then the next interaction begins.


Based on the above settings, we further consider the sparsity problem of conversion signals. We propose a recommendation framework based on deep hierarchical reinforcement learning, including a high-level agent(HRA) and a low-level agent(LRA). 
Both two agents have adapted Actor-Critic architectures.
In order to express our ideas clearly, firstly we define the notations required.
\begin{itemize}
\item {\textbf{High-level state space $S^H$}}: A high-level state $s^h\in S^H$ is defined as user's current long-term preference, which is generated based on user's click and conversion histories, i.e., the items that a user clicked or ordered recently.
\item {\textbf{Low-level state space $S^L$}}: A low-level state $s^l\in S^L$ is defined as user's current short-term preference, which is generated based on user's browsing and click histories, i.e., the items that a user browsed or clicked recently.
\item{\textbf{Goal space $G$}}: A goal $g\in G$ is a signal generated based on current high-level state $s^h$ by HRA and is conveyed to LRA to guide its behavior.
\item{\textbf{Action space $A$}}: An action $a\in A$ is a actual recommendation item generated by LRA based on current low-level state $s^l$.
\item{\textbf{Internal Reward $R^{IN}$}}: After the LRA receives a goal $g$ from HRA, and then takes an action $a$, the LRA receives internal reward $r^{in}(g,a)$. The internal reward is used to evaluate whether the LRA's action follows the goal well.
\item{\textbf{External Reward $R^{EX}$}}: After the LRA takes an action $a$ at the low-level state $s^l$, i.e., recommending an item to a user, the user browses the item and provides his feedback. He can skip, click or order this item, and the LRA receives immediate external reward $r^{ex}(s^{l},a)$ according to the user's feedback.
\item{\textbf{High-level Transition $P^H$}}: High-level transition $p({s^h}'|s^h, g)$ defines the high-level state transition from $s^h$ to ${s^h}'$ when HRA takes goal $g$.
\item{\textbf{High-level Transition $P^L$}}: Low-level transition $p({s^l}'|s^l, a)$ defines the low-level state transition from $s^l$ to ${s^l}'$ when LRA takes action $a$.
\item{\textbf{Discount factor $\gamma$}}:$\gamma\in [0,1]$ defines the discount factor when we measure the present value of future reward. In particular, when $\gamma = 0$, the agents only consider the immediate reward. In other words, when $\gamma = 1$, all future rewards can be counted fully into that of the current action. 
\end{itemize}

Specifically, we model the recommendation task as a MDP in which the recommendation system(including HRA and LRA) interacts with environment $\mathcal{E}$ (or users) over a sequence of time steps. 
The HRA operates at lower temporal resolution and sets abstract goals which are conveyed and enacted by the LRA. 
The LRA generates primitive actions at each time step. 

\begin{figure}[t]
  \centering
  \includegraphics[width=\linewidth,height=4cm]{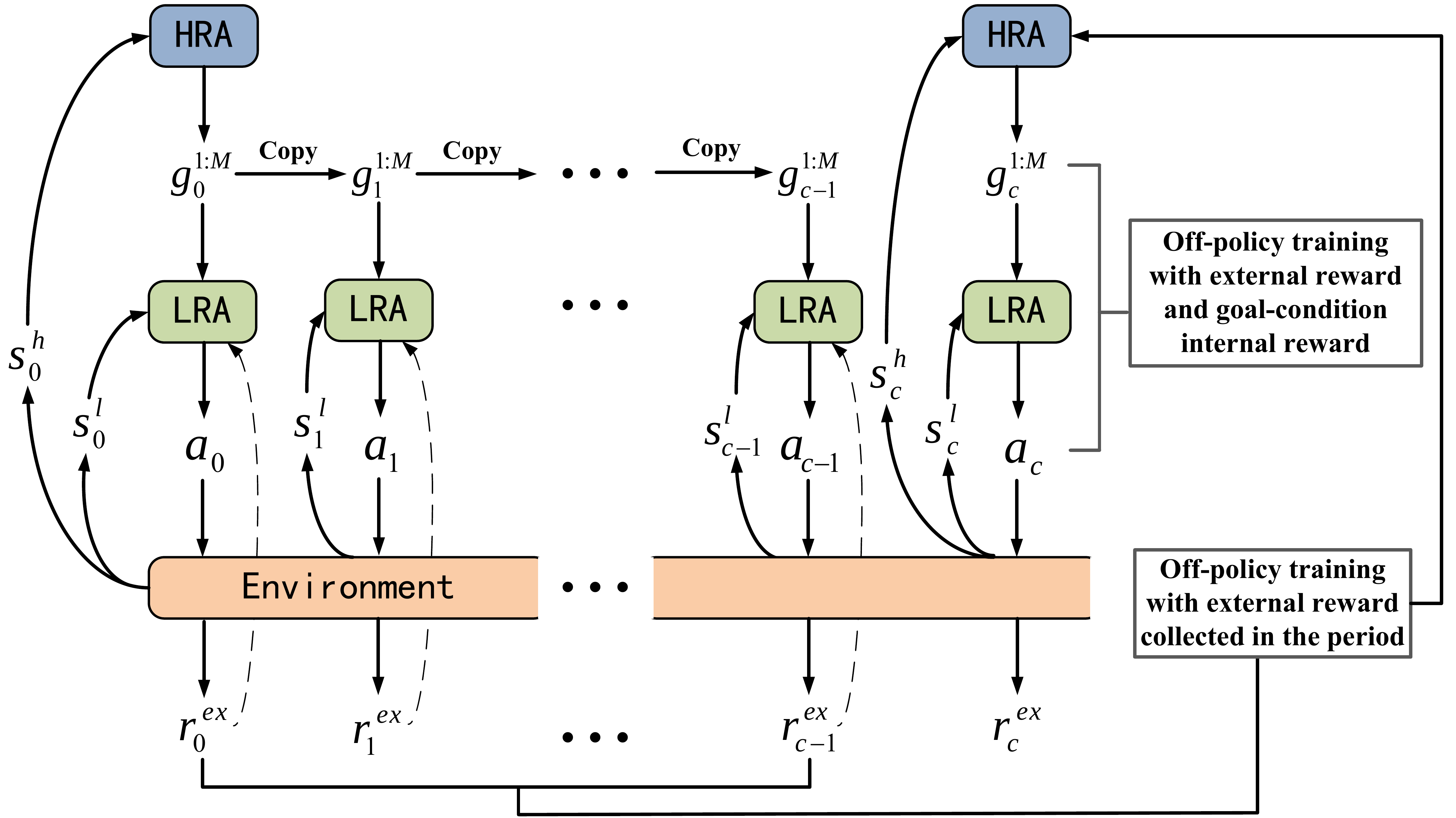}
  \caption{The interaction procedure.}
  \label{imageinteraction}
\end{figure}

As shown in Figure \ref{imageinteraction}, the environment provides a high-level observation state $s^h_t$ and a low-level observation state $s^l_t$ at each time step $t$. 
The HRA observes the high-level state $s^h_t$ and produces a set of goals $g^{1:M}_t=\{g^1_t, g^2_t, \cdots, g^M_t\}$ when $t\equiv 0(mod~c)$. 
This provides temporal abstraction, since HRA produces goals only every $c$ steps, i.e., these goals will be used to guide LRA in the entire $c$ steps. 
The LRA observes the low-level state $s^l_t$ and the set of goals $g^{1:M}_t$, and produces a low-level atomic action $a_t$ based on $s^l_t$, which is applied to the environment. 
Then the LRA receives an internal reward $r^{in}_t$ and a external reward $r^{ex}_t$. 
The internal reward is sampled from the internal reward function $r^{in}_t(g^{1:M}_t, a_t)$, which indicates how the LRA follows the goals.
The external reward is provided by the environment which represents users' actual feedback.
As the consequence of action $a_t$, the environment $\mathcal{E}$ updates the high-level state to $s^h_{t+1}$ with high-level transition $p(s^h_{t+1}|s^h_t,a_t)$ and updates the low-level state to $s^l_{t+1}$ with low-level transition $p(s^l_{t+1}|s^l_t,a_t)$.
After $c$ time steps(from $t$ to $t+c$), the LRA collects recent $c$ external rewards $r^{ex}_{t:t+c-1}=(r^{ex}_t,r^{ex}_{t+1},\cdots,r^{ex}_{t+c-1})$
and conveys them to the HRA to improve its performance.

In the interaction precedure mentioned above, the LRA will store the experience $(s^l_t,g^{1:M}_t,a_t,r^{ex}_t,s^l_{t+1})$ for off-policy training. 
While the HRA will store the experience $(s^h_t,g^{1:M}_t,r^{ex}_{t:t+c-1},s^h_{t+c})$ for off-policy training. 
The goal of hierarchical reinforcement learning is to find a high-level policy $\pi^h:S \rightarrow G$ and a low-level policy $\pi^l:S \rightarrow A$, which can maximize the cumulative external rewards for the recommendation system. 

Both HRA and LRA have adapted Actor-Critic architectures.
The Actor architecture of HRA inputs a high-level state $s^h$ and aims to produce a set of abstract goals $g^{1:M}$.
The Critic architecture inputs the state $s^h$ and the set of goals $g^{1:M}$, and try to evaluate the expected return achievable by the high-level policy as follows:
\begin{equation}
Q^h_i(s^h, g^i) = E_{{s^h}'}\big[r^{high}_i+\gamma Q^h_i({s^h}',{g^i}')|s^h, g^i\big],
\end{equation}
with $1\leq i \leq M$. 
All $Q^h_i, 1\leq i \leq M$ share the same high-level state $s^h$ and evaluate different Q-values of different state-goal pairs.
And $r^{high}_i = \phi_i(r^{ex}_{t:t+c-1})$ represents the reward obtained under goal $g^i$'s guidance.
The Actor architecture of LRA inputs a low-level state $s^l$ and aims to output a deterministic action $a$.
The Critic architecture of LRA inputs this state-action pair $(s^l,a)$, and try to evaluate the expected return achievable by the low-level policy as follows:
\begin{equation}
Q^l(s^l,a)=E_{{s^l}'}\big[r^{low}+\gamma Q^l({s^l}',a')|s^l,a\big],
\end{equation}
where
\begin{equation}
r^{low}=r^{ex}+\alpha r^{in}(g^{1:M},a).    
\end{equation}
represents the total reward that the LRA receives after takes action $a$. And the hyper-parameter $\alpha$ regulates the influence of the internal reward.

Next we will elaborate the HRA and LRA architecture for the proposed framework.
\subsection{Architecture of High-Level Agent}
The high-level agent HRA is designed to generate a set of abstract goals according to user's long-term preference, thus we propose an adapted Actor-Critic architecture for HRA.
We will introduce the encoder structure which is used commonly, and then describe the Actor and Critic architecture of HRA in details.

\subsubsection{\textbf{Encoder for High-Level State Generation}}
We introduce a RNN with Gated Recurrent Units(GRU) to capture users' sequential behaviors as users' long-term preference.
The inputs of GRU are user's last clicked items $\{e^c_1, e^c_2, \cdots, e^c_N\}$ or last ordered items $\{e^o_1, e^o_2, \cdots, e^o_N\}$ (sorted in chronological order) before the current time step, While the output is the representation of users' long-term preference by a vector.
The input $\{e^c_1, e^c_2, \cdots, e^c_N\}$ or $\{e^o_1, e^o_2, \cdots, e^o_N\}$ is dense and low-dimensional vector representations of items.

We leverage GRU rather than Long Short-Term Memory(LSTM) because that GRU outperforms LSTM for capturing users' sequential preference in the recommendation task\cite{Hidasi2015}. 
We use the final hidden state $h_N$ as the output of the RNN layer.
In our framework, two such RNN with GRU are used seperately. 
One of them receives user's last clicked items $\{e^c_1, e^c_2, \cdots, e^c_N\}$ as input and outputs the final hidden state $h^c_N$, while the other one receives user's last ordered items $\{e^o_1, e^o_2, \cdots, e^o_N\}$ as input and outputs the final hidden state $h^o_N$. 
Finally, a linear layer is used to merge the two states and produce the user's long-term preferences:
\begin{equation}
s^h=w_{hc} h^c_N + w_{ho} h^o_N + b_{hs}. 
\label{equstatemerge}
\end{equation}
\subsubsection{\textbf{Actor Framework of HRA}}
\begin{figure}
  \centering
  \includegraphics[width=\linewidth,height=4cm]{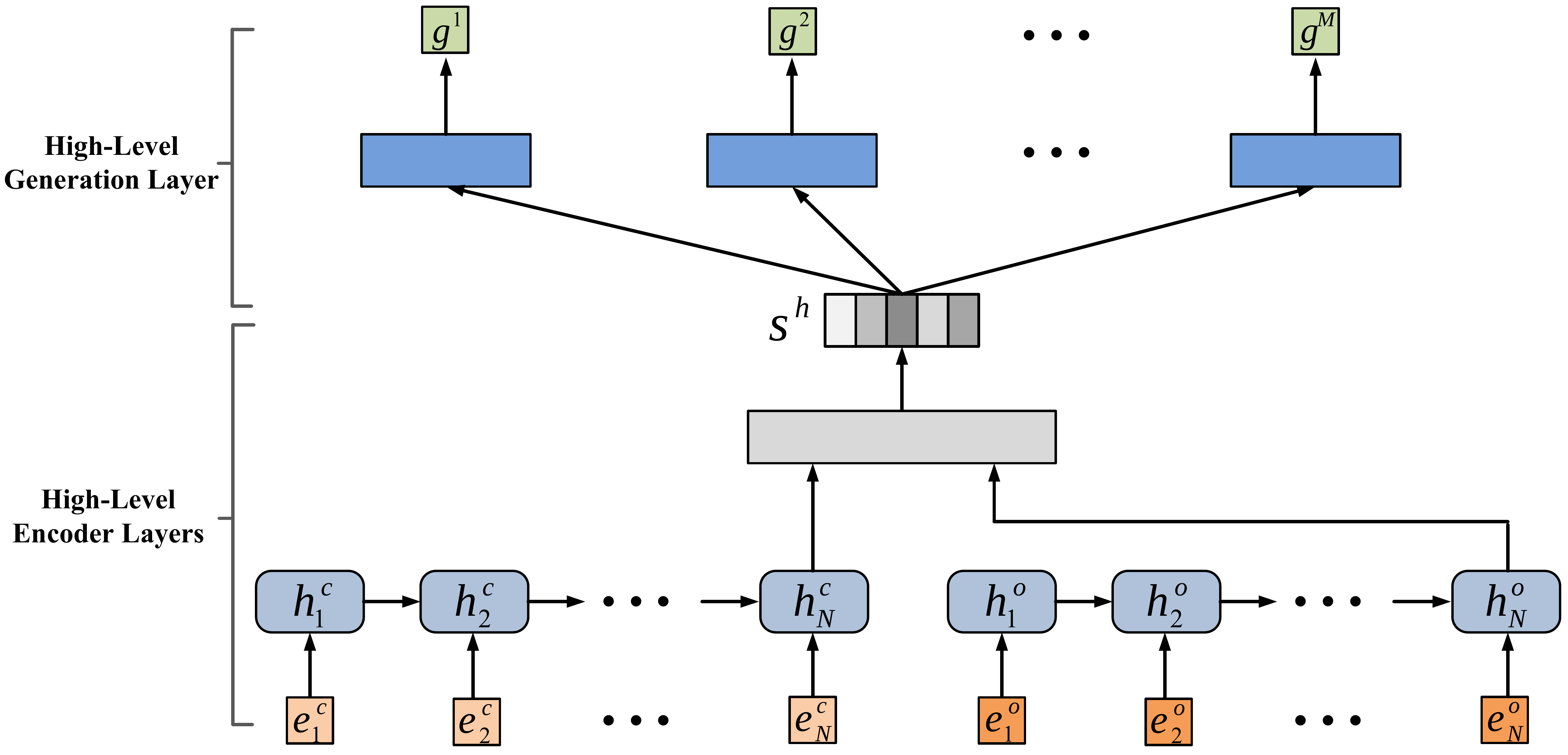}
  \caption{The architecture of high-level actor.}
  \label{imagehactor}
\end{figure}

The Actor framework of HRA, donated by HActor (shown in Figure \ref{imagehactor}), is used to generate multi-goals abstraction based on high-level state $s^h$. 
Thus the encoder structure mentioned above is used firstly to generate the abstract high-level state $s^h$.
Next, in the framework of HActor, $M$ parallel separated fully connected layers are used behind the encoder layers as the goals' generation layer:
\begin{equation}
g^i=B\tanh(w_g^i s^h + b_g^i), 1\leq i \leq M,  
\end{equation}
where parameter $B$ represents the bound of the goals and "tanh" activate function is used since $g_i\in (-B,B)$.

In the framework of HActor, All $M$ goals share the same encoder structure, but their generation layers are different.
That means they get information from the same long-term preference, generate a set of different goals to guide different stages, and improve the encoder and generation layers according to their different feedback.

Due to the existence of the sharing mechanism, in the learning procedure, when each goal gets feedback and updates its related parameters, its generation layer and encoder layers will be updated once. 
Then, the update frequency of parameters in the encoder layers is $M$ times than that in the generation layers.
That has two advantages: 1) the update frequency of HActor is greatly improved; 2) the HActor can obtain information from multiple perspectives, which improves its stability.
\subsubsection{\textbf{Critic Framework of HRA}}
\begin{figure}
  \centering
  \includegraphics[width=\linewidth,height=4.5cm]{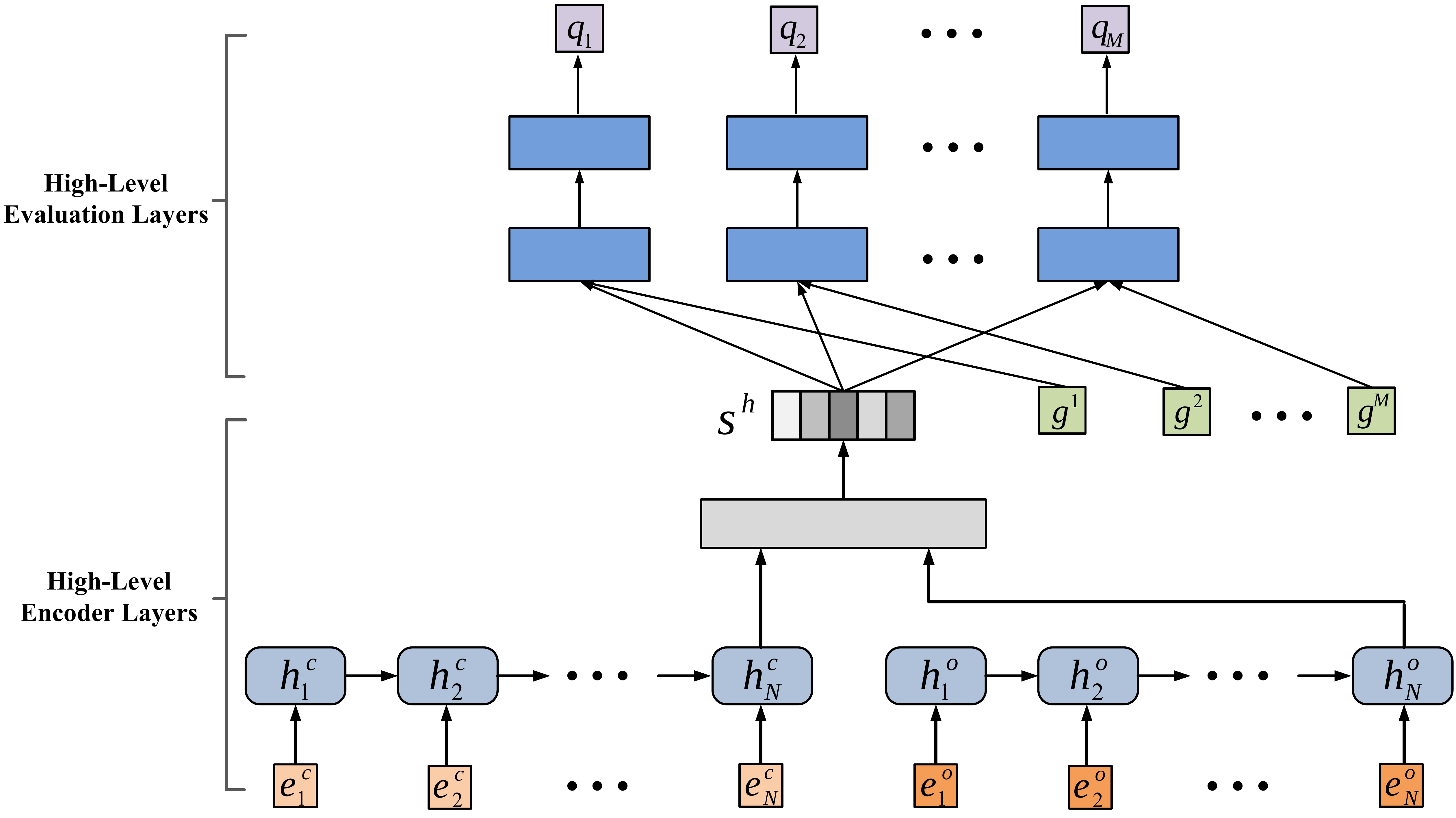}
  \caption{The architecture of high-level critic.}
  \label{imagehcritic}
\end{figure}

The Critic framework of HRA, donated by HCritic (shown in Figure \ref{imagehcritic}), is designed to leverage an approximator to learn multiple goal-value functions $Q^h_i(s^h,g^i), 1\leq i \leq M$, which is a judgment of whether the goals generated by HActor match the current high-level state $s^h$. Then, according to $Q^h_i(s^h,g^i)$, the HActor updates its' related parameters in a direction of improving performance to generate proper goals in the following iterations.
Thus we need to feed user's current high-level state $s^h$ and a set of goals $g^{1:M}$ into the HCritic. 
The same strategy in Eq.(\ref{equstatemerge}) is followed to capture user's long-term preference. 
And then, for each $g^i$, there are 2 fully connected layers used behind the encoder layers as the state-goal pair's evaluation layers:
\begin{equation}
\hat{q_i} = Relu (w^i_{hcs}s^h + w^i_{hcg}g^i+ b^i_{hc}),
\end{equation}
\begin{equation}
q_i = Relu (w^i_{hq}\hat{q_i}+b^i_{hq}),
\end{equation}
where $1\leq i \leq M$ and we use the activation function "Relu" since $q_i \in (0,+\infty)$. 

In the framework of HCritic, $M$ parallel separated evaluation layers are placed behind the same encoder layers, estimating the expected returns of the goals according to their benefit assignment functions $\phi_i(r^{ex}_{t:t+c-1})$.
The benefit assignment function is mainly related to the rewards in the stage in which the goal is used , and the compensation when the low-level strategy has not converged is also considered. 
We will discuss the benefit assignment function in Section 3 with more details. 
Similarly, due to the sharing mechanism, the update speed and convergence stability of HCritic are also improved.
\subsection{Architecture of Low-Level Agent}
The low-level agent LRA is designed to generate a set of actual recommendation items according to user's short-term preference, thus we propose an adapted Actor-Critic architecture for LRA.
We will introduce the encoder structure which is used commonly, and then describe the Actor and Critic architecture of LRA in details.

\subsubsection{\textbf{Encoder for Low-Level State Generation}}
In our framework, two RNN with GRU similar to that mentioned in Section 2.2.1 are used seperately.
One of them receives user's last browsed items $\{e^b_1, e^b_2, \cdots, e^b_N\}$ as input and outputs the final hidden state $h^b_N$, while the other one receives user's last clicked items $\{e^c_1, e^c_2, \cdots, e^c_N\}$ as input and outputs the final hidden state $h^c_N$. 
Finally, a linear layer is used to merge the two states and produce the user's short-term preferences:
\begin{equation}
s^l=w_{lb} h^b_N + w_{lc} h^c_N + b_{ls}.    
\end{equation}
\subsubsection{\textbf{Actor Framework of LRA}}
\begin{figure}[t]
  \centering
  \includegraphics[width=\linewidth,height=4.5cm]{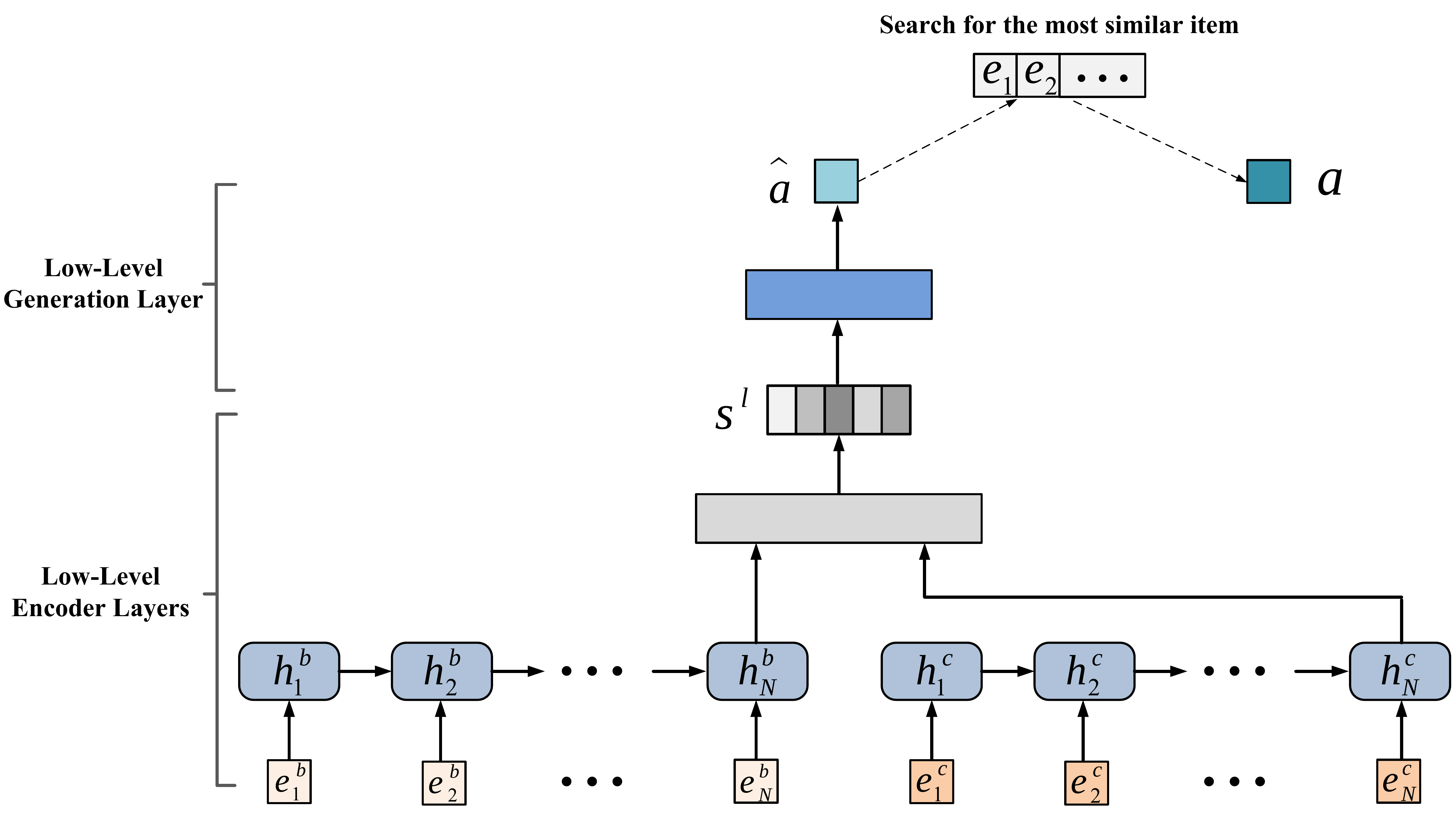}
  \caption{The architecture of low-level actor.}
  \label{imagelactor}
\end{figure}

The Actor framework of LRA, donated by LActor (shown in Figure \ref{imagelactor}), is used to generate actual recommendation items based on low-level state $s^l$. 
Thus the encoder structure mentioned above is used firstly to generate the abstract low-level state $s^l$.
Next, in the framework of LActor, a fully connected layer is used behind the encoder layers as the action generation layer:
\begin{equation}
\hat{a}=B\tanh(w_a s^l + b_a),    
\end{equation}
where parameter $B$ represents the bound of the action and "tanh" activate function is used since $a\in (-B,B)$.

Notice that the generated item embedding $\hat{a}$ may be not in the real item embedding set, that we need to map it to valid item embedding, which will be provided in Section 3.
\subsubsection{\textbf{Critic Framework of LRA}}
\begin{figure}[t]
  \centering
  \includegraphics[width=\linewidth,height=4.5cm]{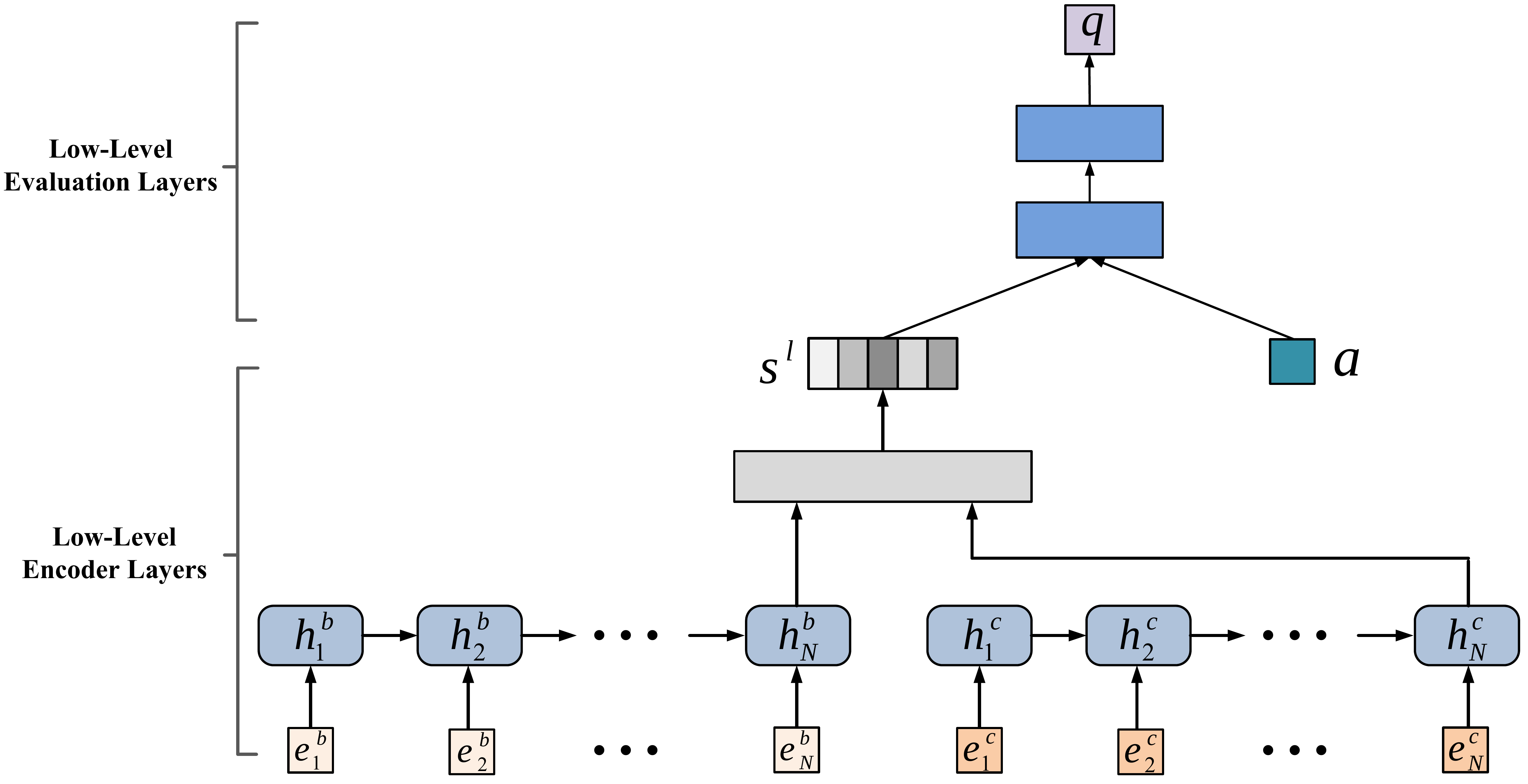}
  \caption{The architecture of low-level critic.}
  \label{imagelcritic}
\end{figure}

The Critic framework of LRA, donated by LCritic (shown in Figure \ref{imagelcritic}), is designed to leverage an approximator to learn action value functions $Q^l(s^l,a)$, which is a judgment of whether the action generated by LActor matches the current low-level state $s^l$ and follows the guidance of the goals well. Then, according to $Q^l(s^l,a)$, the LActor updates its' parameters in a direction of improving performance to generate proper actions in the following iterations.

Thus we need to feed user's current low-level state $s^l$ and action $a$ into the LCritic. 
The same encoder layers as LActor's are used to capture user's short-term preference. 
And then, there are 2 fully connected layers used behind the encoder layers as the state-action pair's evaluation layers:
\begin{equation}
\hat{q} = Relu (w_{lcs}s^l + w_{lca}a+ b_{lc}),
\end{equation}
\begin{equation}
q = Relu (w_{lq}\hat{q}+b_{lq}),
\end{equation}
where the activation function "Relu" is used since $q \in (0,+\infty)$. 

As mentioned in Eq. (2)(3), the update direction of LCritic will be affected by both the external reward $r^{ex}$ and the internal reward $r^{in}$. 
The form of the internal reward function  $r^{in}(g^{1:M}, a)$ determines the way the goals guide the LRA. 
Thus a reasonable internal reward function is needed to make goals play different roles at different stages.
In this work, we cut a period of $c$ steps into average $M$ parts, and use each goal in each $\lfloor c/M \rfloor$ steps.
Cosine similarity is used to measure the gap between action and corresponding goal and produce the internal reward:
\begin{equation}
r^{in}_t(g^{1:M}_t, a_t) = \frac{a_t^T\cdot g^j_t}{\left\| a_t \right\| \left \|g^j_t \right\|},\lfloor c/M \rfloor(j-1) < t \leq \lfloor c/M \rfloor j.
\end{equation}

Notice that only one goal $g^j$ is used in each time step, and the internal reward function can be  simplified as $r^{in}_t(g^j_t, a_t)$. 
The responsibility of each goal is clearly defined and the time consumption is reduced.
Other reasonable designs that promote the diversity function of the goals are also encouraged.

\section{Training procedure}
With the proposed recommendation framework based on hierarchical reinforcement learning, we will discuss the work of parameter training in this section. 
We first propose an online training algorithm, and then test the framework in the online environment and offline history logs respectively. The details of the test procedure are shown in Appendix C.

\subsection{Actual Action Mapping}
As mentioned in section 3.3.2, we generate a recommendation item embedding $\hat{a}$ using the user's short-term preferences $s^l$. But $\hat{a}$ is a virtual-action because it may not be in the real item embedding set $I$. So we have to map this virtual-action  $\hat{a}$ into a real action $a$ (a real item embedding). Under this setting, for each $\hat{a}$, we choose the most similar $a \in I$ as the real item embedding. In this work, we use cosine similarity as the metric:
\begin{equation}
a=\arg \max_{a_i\in I} \frac{\hat{a}^T \cdot a_i}{\left\| \hat{a} \right\| \left \| a_i \right\|} = \arg \max_{a_i\in I} \hat{a}^T \cdot \frac{a_i}{\left \| a_i \right\|}.
\end{equation}
To reduce the amount of computation, we pre-compute $\frac{a_i}{\left \| a_i \right\|}$ for all $a_i\in I$ and use the item recall mechanism to eliminate irrelevant and redundant items. The details of Mapping Algorithm are shown in Appendix B.1.

Note that when the item embedding set $I$ is large, the above method faces the challenge of insufficient computation time and storage space. A nearest neighbor search method based on Hash mapping can map high-dimensional data into a series of compact binary codes\cite{hash2011}.
And the similarity relation between the original high-dimensional data is approximated by the distance between the binary codes. It can achieve high calculation speed and reduce storage consumption at the expense of acceptable error, which can be used as an alternative of Algorithm 1.

\subsection{Benefit Assignment Function}
As mentioned in Section 3.2.2, the benefit assignment function $\phi_i(r^{ex}_{t:t+c-1}),1\leq i\leq M$ assigns the external reward of the recent c-steps $r^{ex}_{t:t+c-1}$ collected by LRA to each goal. There are two main factors to consider: 1. How does LRA perform under the guidance of each goal? 2. How to coordinate different goals in a consistent direction?

In Section 3.3.3, when determine the way the goal guide LRA, we cut a period of $c$ steps into average $M$ parts, and use each goal in each $\lfloor c/M \rfloor$ steps. 
Thus a natural idea is that we collect the rewards in each $\lfloor c/M \rfloor$ steps and assign them to the corresponding goal:
\begin{equation}
\phi_i^0(r^{ex}_{t:t+c-1})=\sum^{\lfloor c/M \rfloor i-1}_{j=\lfloor c/M \rfloor(i-1)} r^{ex}_{t+j}.
\end{equation}

However, this assignment method does not consider the latter factor. To deal with this problem, we propose an extended benefit assignment function based on the Eq.(14):
\begin{equation}
\phi'_i(r^{ex}_{t:t+c-1})=\sum^{i}_{k=1}\beta^{i-k}\phi_k^0(r^{ex}_{t:t+c-1}) ,
\end{equation}
where parameter $\beta$ is the high-level benefit discount factor.
In Eq.(15), each goal is assigned with the cumulative discounted external rewards from the beginning of current period of $c$ steps to the stage in which it is used, forcing the subsequent goals to improve the overall performance of the entire period. When $\beta=0$, it is equivalent to Eq.(14); when $\beta=1$, all related reward should be considered equally. 
\subsection{Training Algorithm}
In the proposed recommendation framework based on hierarchical reinforcement learning, both the high-level agent and the low-level agent have adapted Actor-Critic architectures. We utilize DDPG algorithm to train the parameters of both agents. The details of the Online Training Algorithm are shown in Appendix B.2.

In the high-level agent HRA, the HCritic can be trained by minimizing a series of loss function $L(\Theta_{\mu_i}^h),1\leq i \leq M$ as:
\begin{equation}
\begin{split}
L(\Theta_{\mu_i}^h)=E_{s^h,g^i,r^{high}_i,{s^h}'}\big[\big(r^{high}_i+\gamma Q_{{\Theta_{\mu_i}^h}'}\big({s^h}',f_{\Theta_{\pi_i}^h}({s^h}')\big)\\-Q_{\Theta_{\mu_i}^h}(s^h,g^i)\big)^2\big],
\end{split}
\end{equation}
where $\Theta_{\mu_i}^h$ represents all parameters used to generate the Q-value $q^i$, which includes the parameters in the shared encoder layers and the $i$-th evaluation layers of HCritic. The HCritic is trained from samples stored in a high-level replay buffer.

The first term $y_i=r^{high}_i+\gamma Q_{{\Theta_{\mu_i}^h}'}({s^h}',f_{\Theta_{\pi_i}^h}({s^h}'))$ in Eq.(16) is the target for the current period of $c$ steps. The parameters from the previous period ${\Theta_{\mu_i}^h}'$ are fixed when optimizing the loss function $L(\Theta_{\mu_i}^h)$. In practice , it is often computationally efficient to optimize the loss function by stochastic gradient descent, rather than computing the expectations over the experience space. The derivatives of loss function $L (\Theta_{\mu_i}^h)$ with respective to parameters $\Theta_{\mu_i}^h$ are represented as follows:
\begin{equation}
\begin{split}
\bigtriangledown L(\Theta_{\mu_i}^h)=E_{s^h,g^i,r^{high}_i,{s^h}'}\big[\big(r^{high}_i+\gamma Q_{{\Theta_{\mu_i}^h}'}\big({s^h}',f_{\Theta_{\pi_i}^h}({s^h}')\big)\\-Q_{\Theta_{\mu_i}^h}(s^h,g^i)\big)\bigtriangledown_{\Theta_{\mu_i}^h}Q_{\Theta_{\mu_i}^h}(s^h,g^i)\big], 1\leq i \leq M.
\end{split}
\end{equation}

The HActor is updated with the policy gradient:
\begin{equation}
\bigtriangledown_{\Theta_{\pi_i}^h}f_{\Theta_{\pi_i}^h}=E_{s^h}\big[\bigtriangledown_{g^i}Q_{\Theta_{\mu_i}^h}(s^h,g^i)\bigtriangledown_{\Theta_{\pi_i}^h}f_{\Theta_{\pi_i}^h}(s^h)\big], 1\leq i \leq M,
\end{equation}
where $g^i=f_{\Theta_{\pi_i}^h}(s^h)$. 

Similarly, in the low-level agent LRA, the LCritic can be trained by minimizing the loss function $L(\Theta_{\mu}^l)$ as:
\begin{equation}
L(\Theta_{\mu}^l)=E_{s^l,a,r^{low},{s^l}'}\big[\big(r^{low}+\gamma Q_{{\Theta_{\mu}^l}'}\big({s^l}',f_{\Theta_{\pi}^l}({s^l}')\big)-Q_{\Theta_{\mu}^l}(s^l,a)\big)^2\big],  
\end{equation}
where $\Theta_{\mu}^l$ represents all parameters in LCritic. The LCritic is trained from samples stored in a low-level replay buffer. Actions stored in the low-level replay buffer are generated by valid-action $a$. This allows the learning algorithm to leverage the information of which action was actually executed to train the LCritic\cite{Gabriel2015}.
The derivatives of loss function $L (\Theta_{\mu}^l)$ with respective to parameters $\Theta_{\mu}^l$ are represented as follows:
\begin{equation}
\begin{split}
\bigtriangledown L(\Theta_{\mu}^l)=E_{s^l,a,r^{low},{s^l}'}\big[\big(r^{low}+\gamma Q_{{\Theta_{\mu}^l}'}\big({s^l}',f_{\Theta_{\pi}^l}({s^l}')\big)\\-Q_{\Theta_{\mu}^l}(s^l,a)\big)\bigtriangledown_{\Theta_{\mu}^l}Q_{\Theta_{\mu}^l}(s^l,a)\big].
\end{split}
\end{equation}

The LActor is updated with the policy gradient:
\begin{equation}
\bigtriangledown_{\Theta_{\pi}^l}f_{\Theta_{\pi}^l}=E_{s^l}\big[\bigtriangledown_{\hat{a}}Q_{\Theta_{\mu}^l}(s^l,\hat{a})\bigtriangledown_{\Theta_{\pi}^l}f_{\Theta_{\pi}^l}(s^l)\big],
\end{equation}
where $\hat{a}=f_{\Theta_{\pi}^l}(s^l)$, i.e., $\hat{a}$ is generated by virtual-action. Note that virtual-action is the actual output of LActor. This guarantees that policy gradient is taken at the actual output of policy $f_{\Theta_{\pi}^l}$ \cite{Gabriel2015}.

\section{Experiments}
In this session, we conduct extensive experiments with a dataset from a real e-commerce company to evaluate the effectiveness of the proposed framework. We mainly focus on two questions: 1) how the proposed framework performs compared to representative baselines; and 2) how the components in the framework contribute to the performance. We first introduce experimental settings. Then we seek answers to the above two questions. Finally, we discuss the impact of important parameters.
\subsection{Experiment Settings}
We evaluate our method on a dataset of August, 2018 from a real e-commerce company. 
The statistics about the dataset are shown in Appendix D.

We do online training and test on a simulated online environment. The simulated online environment is trained on users' logs. The simulator has the similar architecture with LCritic, while the output layer is a softmax layer that predicts the immediate feedback according to current low-level state $s^l_t$ and a recommendation item $a_t$. We test the simulator on users' logs, and experimental results demonstrate that the simulated online environment has overall 90\% precision for immediate feedback prediction task. This result suggests that the simulator can accurately simulate the real online environment and predict the online rewards, which enables us to train and test our model on it.

For a new session, the initial high-level and low-level state are collected from the previous sessions of the user. In this work, we leverage $N=10$ previously browsed/clicked/ordered items to generate high-level and low-level state. The external reward $r^{ex}$ of skipped/clicked/ordered are empirically set as 0, 1, and 5, respectively. The dimension of the embedding of items is 50, and we set the discounted factor $\gamma=0.95$. For the parameters of the proposed framework, we select them via cross-validation. Corresponding, we also do parameter-tuning for baselines for a fair comparison. 

For online test, we leverage the average summation of all rewards in one recommendation session as the metric. For offline test, we select \textbf{MAP}\cite{map} and \textbf{NDCG@20(40)}\cite{ndcg} as the metrics to measure the performance. The difference of ours from traditional Learn-to-Rank methods is that we rank both clicked and ordered items together,and set them by different rewards, rather than only rank clicked items as that in the Learn-to-Rank setting.

\begin{figure}
\begin{minipage}{0.48\linewidth}
  \centerline{\includegraphics[width=4.8cm]{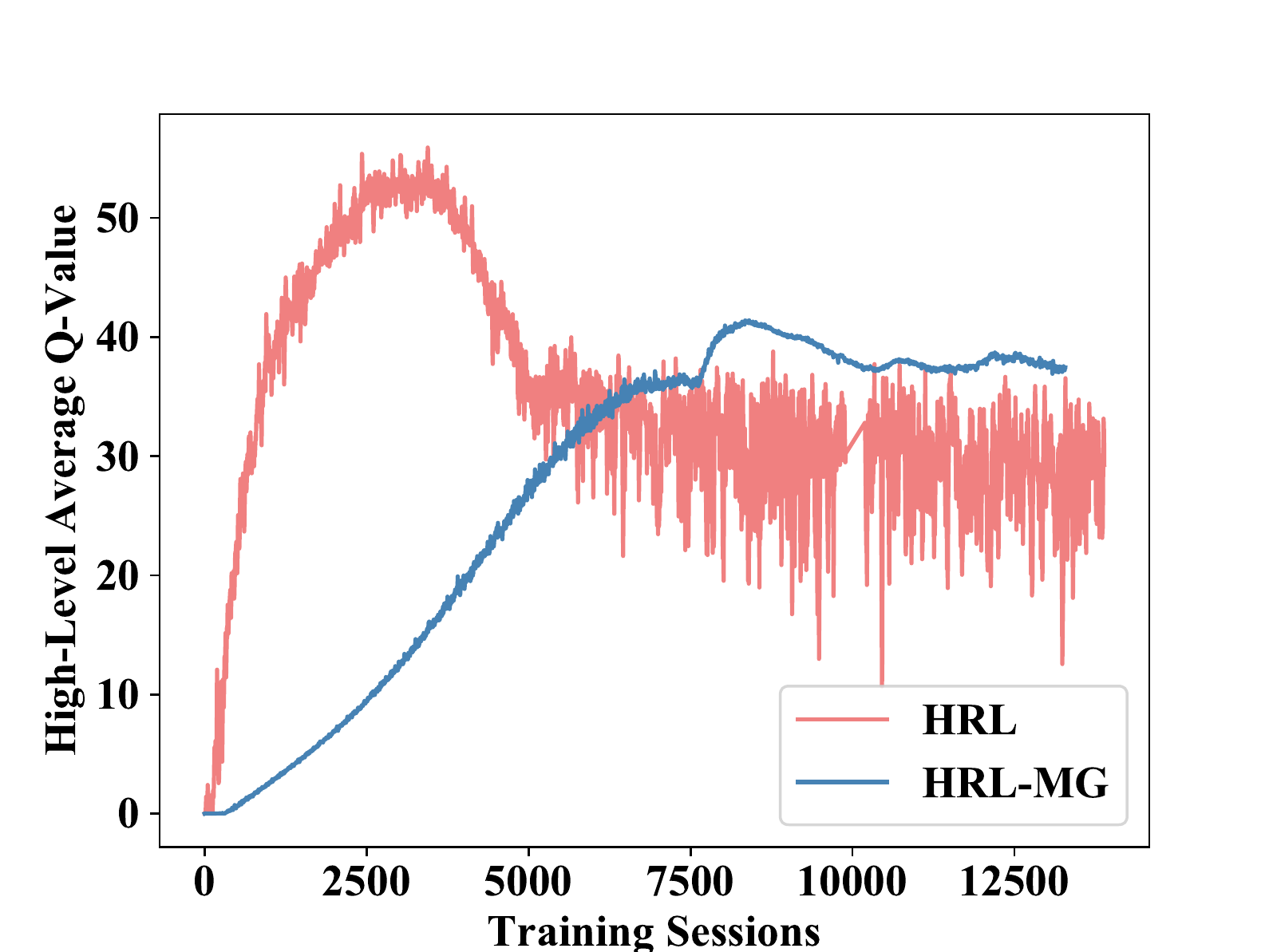}}
  \centerline{(a)}
\end{minipage}
\hfill
\begin{minipage}{.48\linewidth}
  \centerline{\includegraphics[width=4.8cm]{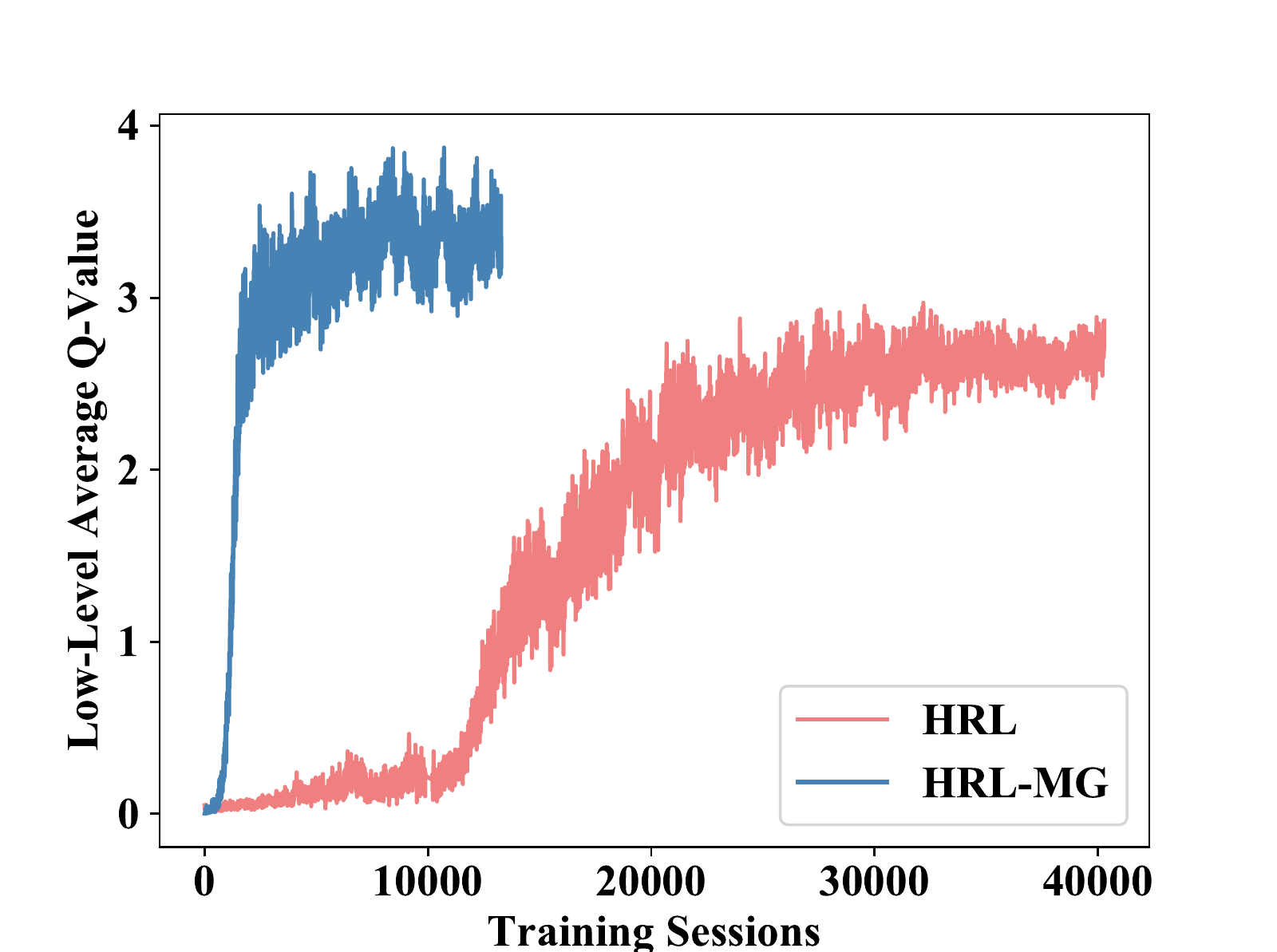}}
  \centerline{(b)}
\end{minipage}
\caption{Training procedure.}
\label{imagetrain}
\end{figure}

\subsection{Performance Comparison}
First we train the proposed framework HRL-MG to converge in the simulated online environment and then test the performance both in online and offline ways, and compare our framework with DNN, DDPG and HRL. 
\begin{itemize}
\item {\textbf{DNN}}: This is a deep neural network similar to LCritic, with similar encoder layers to catch user's abstract state and try to evaluate the immediate reward of the current state-action pair. It always recommends items with highest immediate reward.
\item {\textbf{DDPG}}: Only the low-level agent is used without goals' guidance. It always recommends items with highest accumulated decay returns evaluated by LCritic.
\item{\textbf{HRL}}: The proposed framework with $M$ set to 1. Its high level agent guides the low level agent with only one goal in $c$ time steps.
\end{itemize}

Here we utilize online training strategy to train DDPG and HRL(similar to method mentioned in Section 3.3). DNN is also applicable to be trained via the rewards generated by simulated online environment. 

We do offline test by re-ranking users' offline logs, while do online test on the simulated online environment mentioned above.
As the online test is based on the simulator, we can artificially control the length of recommendation sessions to study the performance in short and long sessions. We define short sessions with 50 recommendation items, while long sessions with 300 recommendation items. The results are shown in Figure \ref{imagetrain}-\ref{imageon}. It can be observed:
\begin{figure}
\begin{minipage}{0.48\linewidth}
  \centerline{\includegraphics[width=4.8cm]{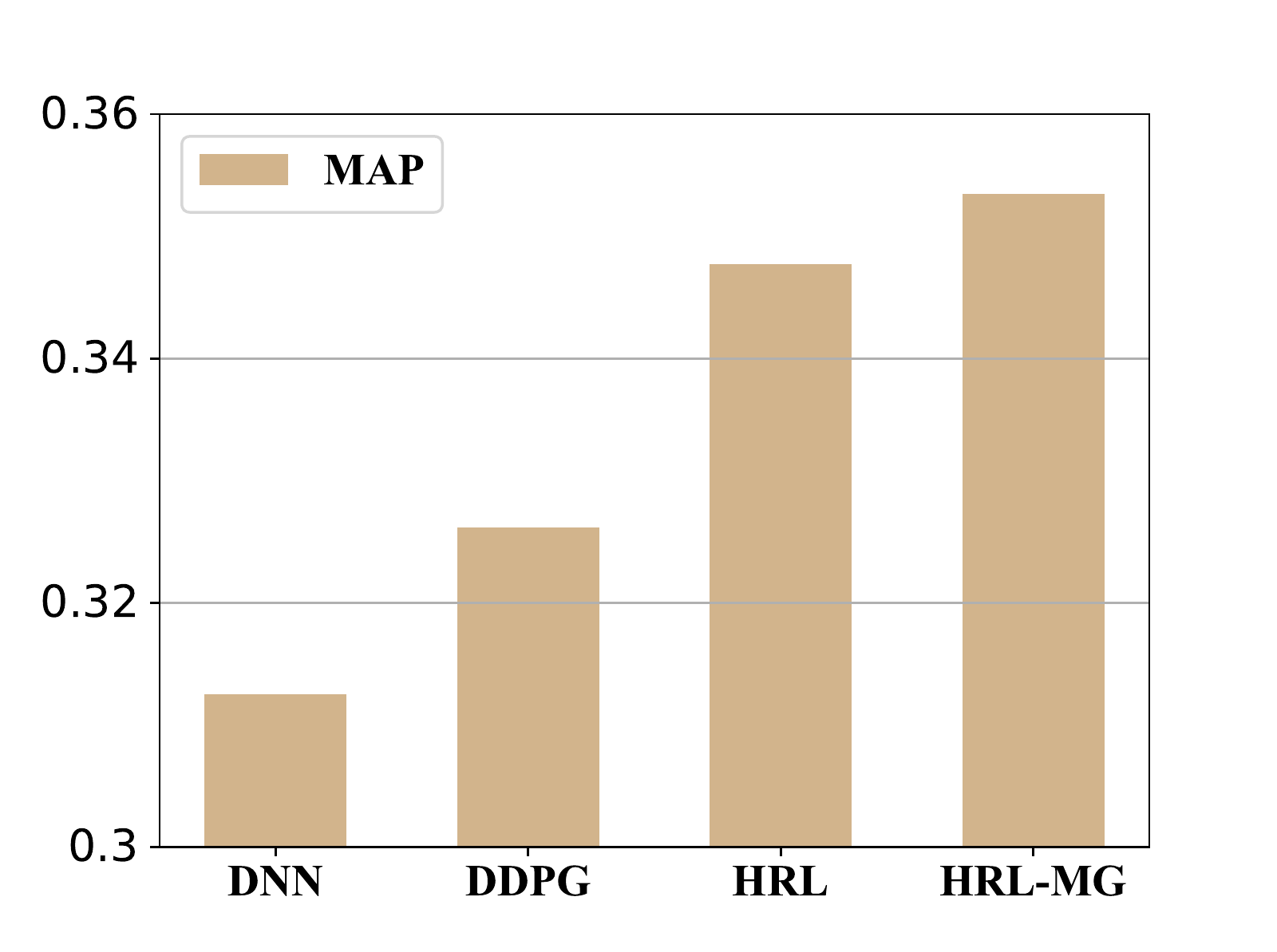}}
  \centerline{(a)}
\end{minipage}
\hfill
\begin{minipage}{.48\linewidth}
  \centerline{\includegraphics[width=4.8cm]{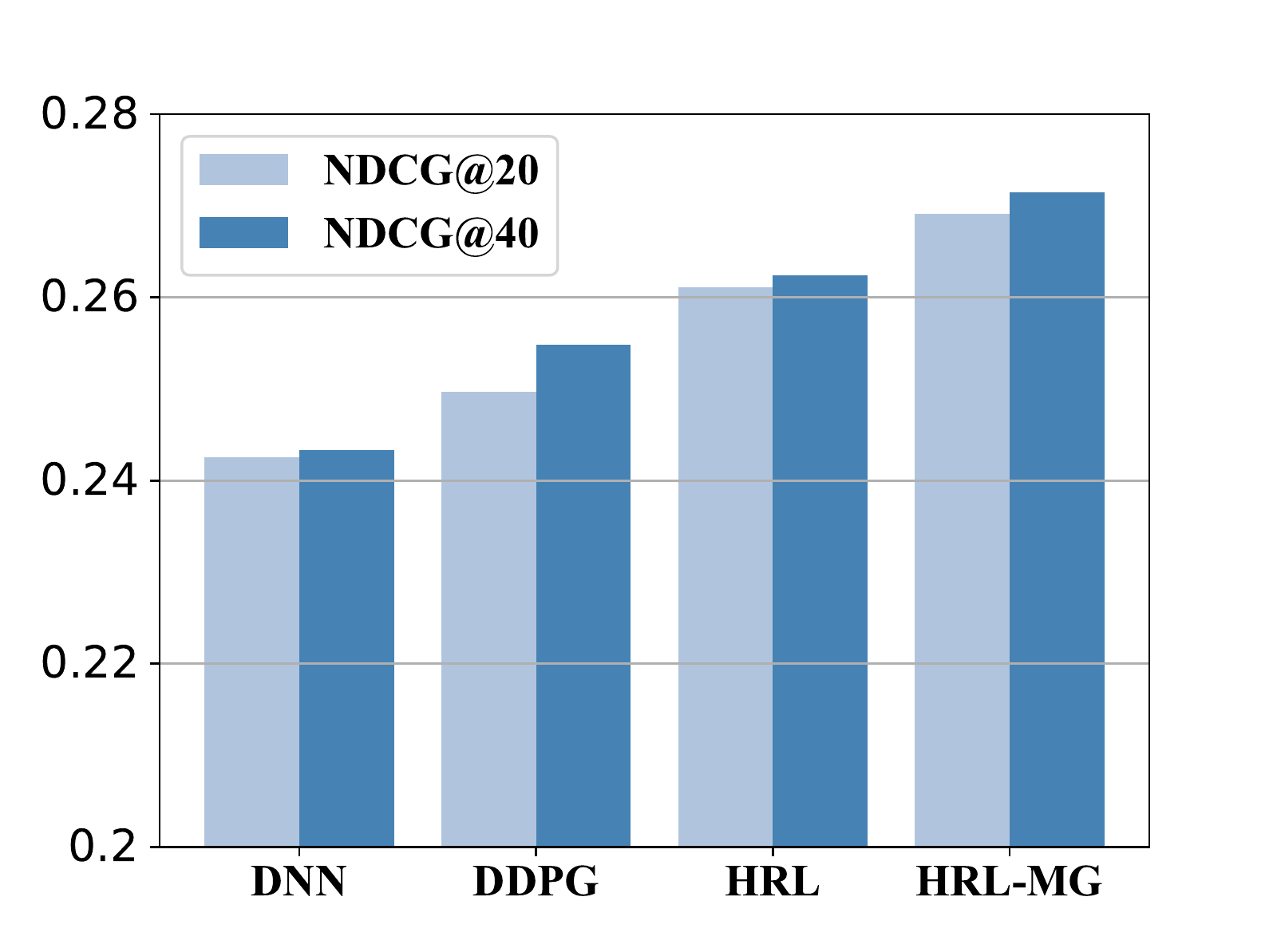}}
  \centerline{(b)}
\end{minipage}
\caption{Performance comparison for offline test.}
\label{imageoff}
\end{figure}
\begin{figure}
\begin{minipage}{0.48\linewidth}
  \centerline{\includegraphics[width=4.8cm]{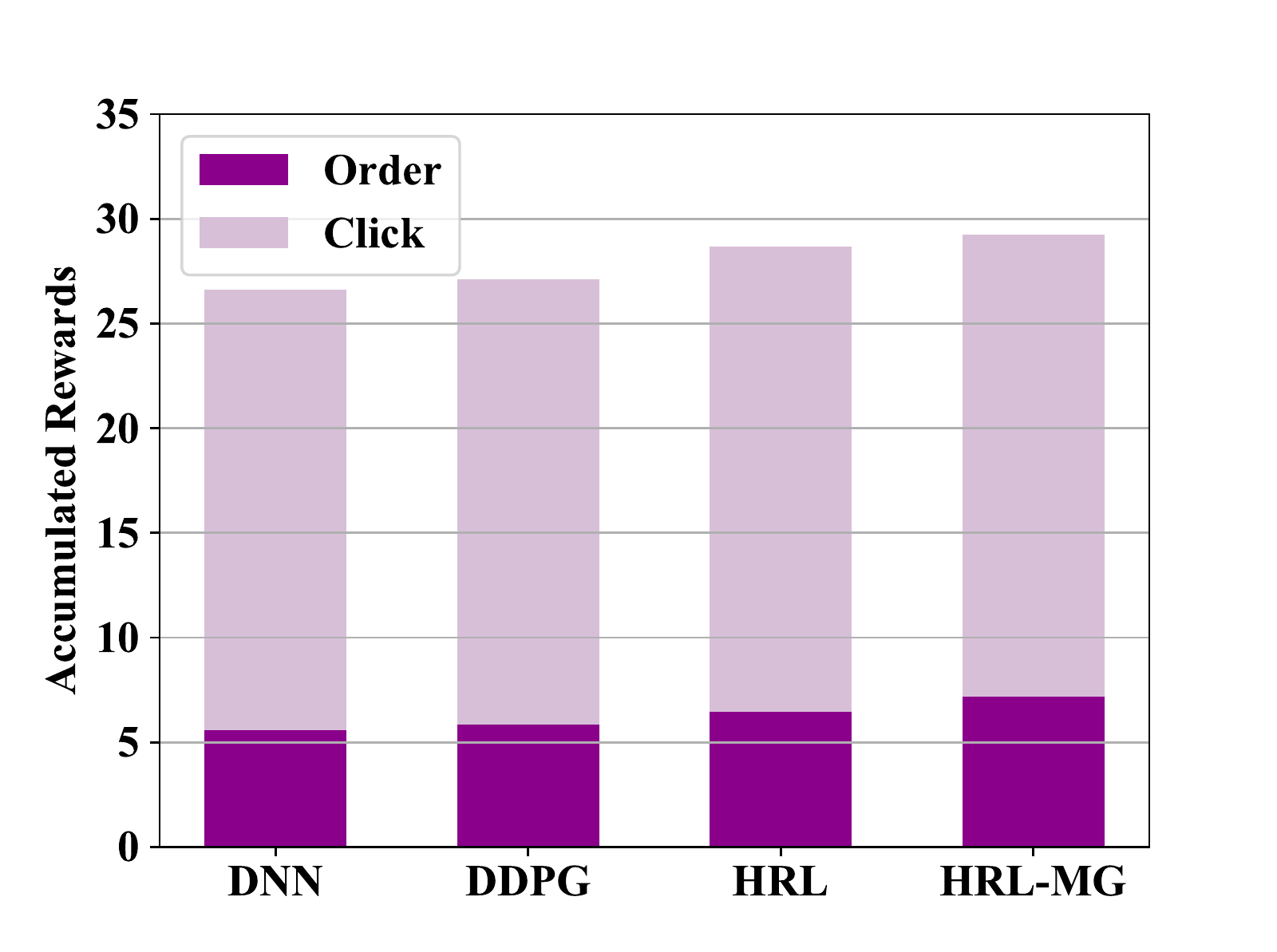}}
  \centerline{(a) Performance in long sessions.}
\end{minipage}
\hfill
\begin{minipage}{0.48\linewidth}
  \centerline{\includegraphics[width=4.8cm]{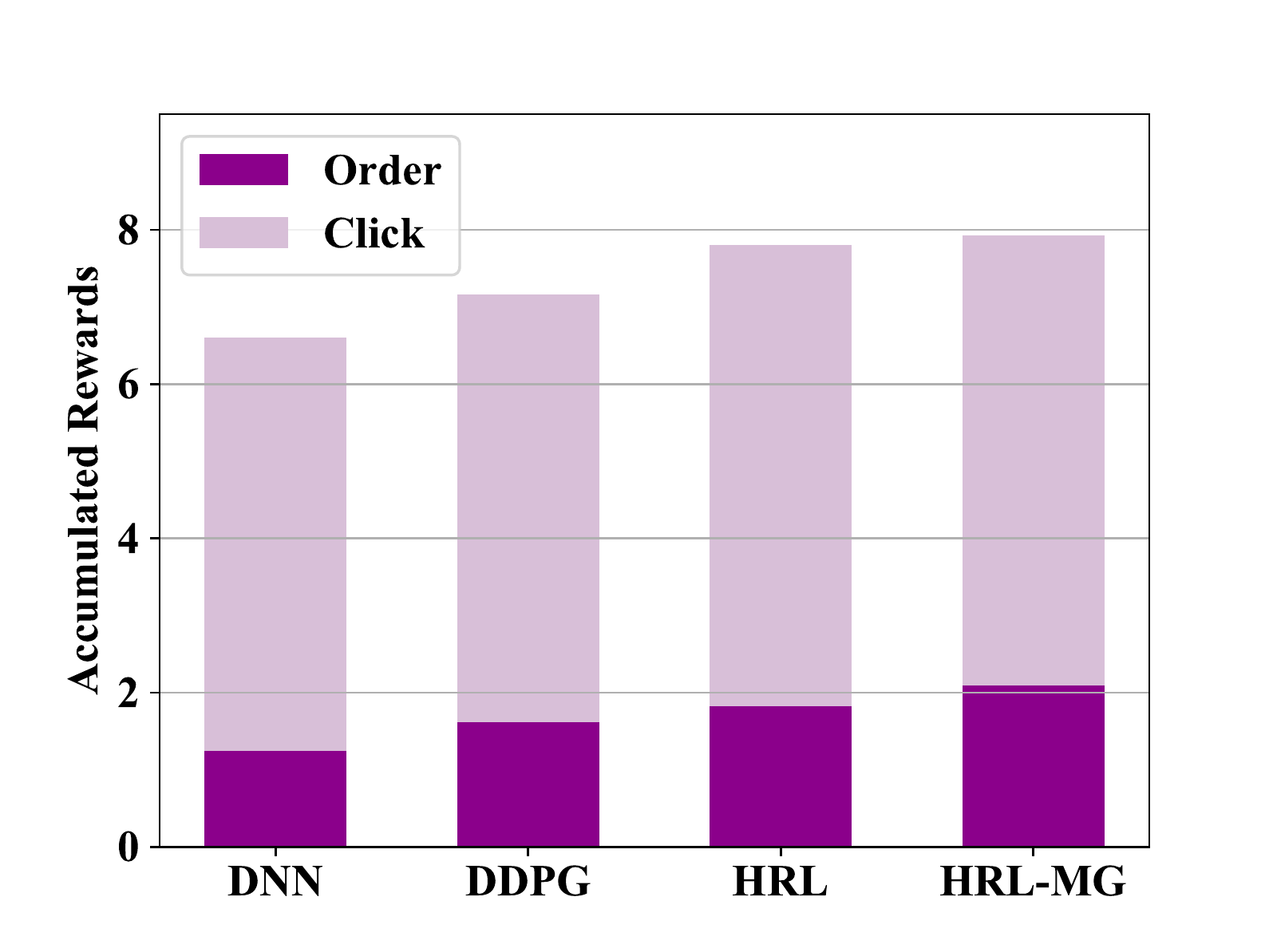}}
  \centerline{(b) Performance in short sessions.}
\end{minipage}
\caption{Performance comparison for online test.}
\label{imageon}
\end{figure}
\begin{itemize}
\item Figure \ref{imagetrain}(a)(b) illustrate the training process of high-level and low-level agent in HRL and our HRL-MG. In Figure 6(a), the high-level agent in HRL has a wrong growth, eventually falls back to the convergence position, while the HRA of HRL-MG grows steadily. This is because multiple goals and sharing mechanism improve the update speed and stability of the high-level agent. In Figure 6(b), the convergence speed of low-level agent in HRL-MG is much faster than that in HRL. Notice that the low-level agent in HRL begins to evolve until the high-level agent converges, while that in HRL-MG doesn't need. This is because multiple goals greatly reduce the difficulty for the low-level agent to achieve the goal.
\item Figure 7,8 show that DDPG, HRL, and HRL-MG are better than DNN both in offline and online test. This is because DNN only considers immediate reward, while the other three are based on the reinforcement learning, taking long-term cumulative returns into account and achieving higher performance.
\item Figure 7,8 show that HRL and HRL-MG are better than DDPG both in offline and online test. This is because DDPG acts at high time resolution, generates each specific recommendation item according to the current state, and cannot effectively handle sparse conversion signal. While HRL and HRL-MG have hierarchical structures, which can observe in a wider time range, capture the sparse reward signal, and improve the performance of the low-level agent through the guidance of the goals.
\item Figure 7 shows that the performance of HRL-MG is better than HRL's in offline test. This is because multi-goals help to convey more sparse conversion information, forcing the low-level agent to focus more on improving the conversion.
\item Figure 8 shows that in online test, the cumulative total rewards and orders of HRL-MG are significantly higher than HRL's, and the cumulative clicks are slightly lower. This shows that HRL-MG is better in improving conversions and overall revenue. There exists a trade-off between click and conversion enhancements because they are not completely positively correlated.
\end{itemize}

\subsection{Parameter Sensitivity}
Our method has two key parameters: $\alpha$ that controls the influence of internal reward and $M$ controls the number of goals.
To study the impact of these parameters, we investigate how the proposed framework works with the changes of one parameter, while fixing other parameters.

\begin{figure}
\begin{minipage}{0.32\linewidth}
  \centerline{\includegraphics[width=2.9cm]{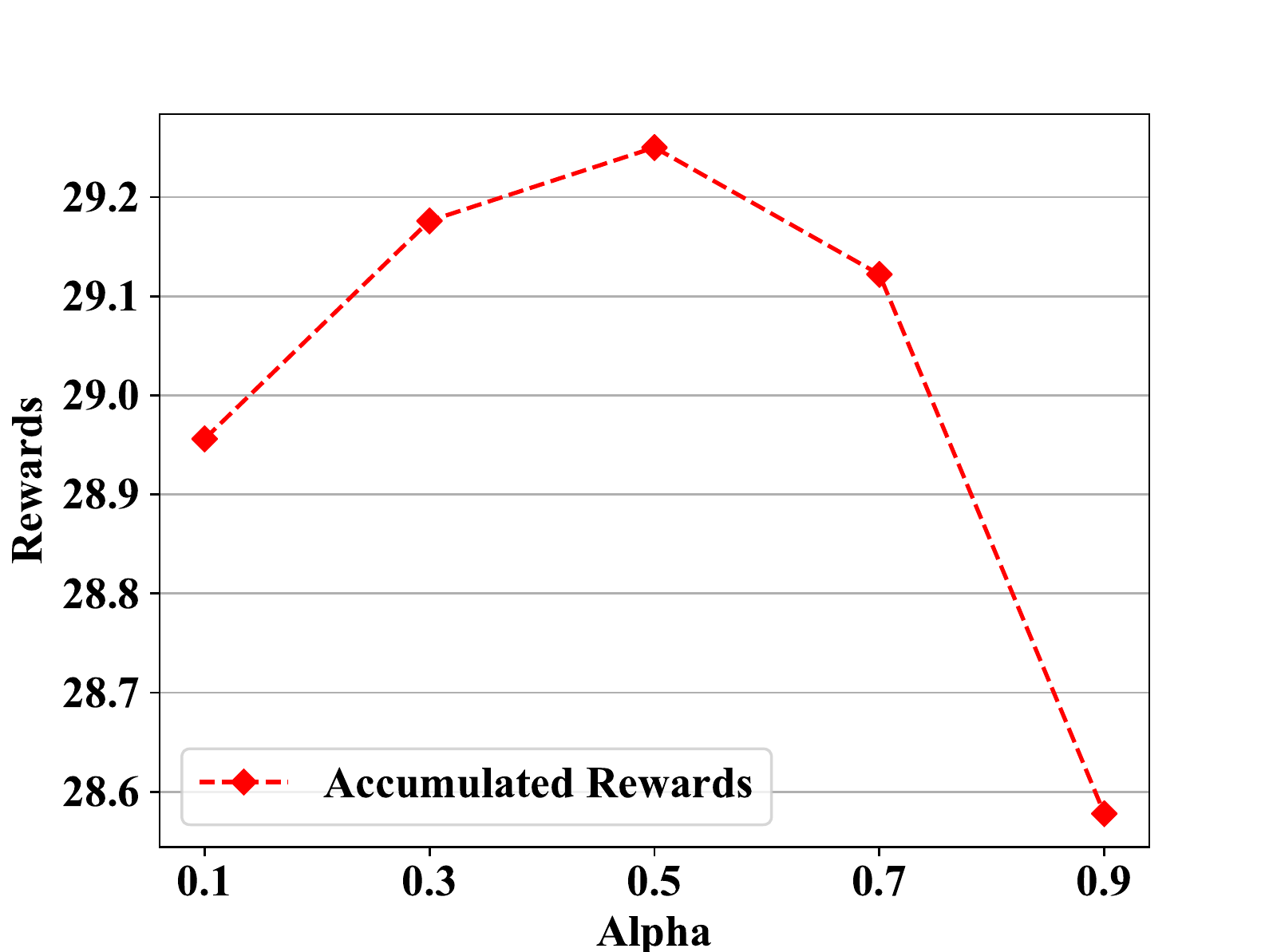}}
  \centerline{(a)}
\end{minipage}
\hfill
\begin{minipage}{0.32\linewidth}
  \centerline{\includegraphics[width=2.9cm]{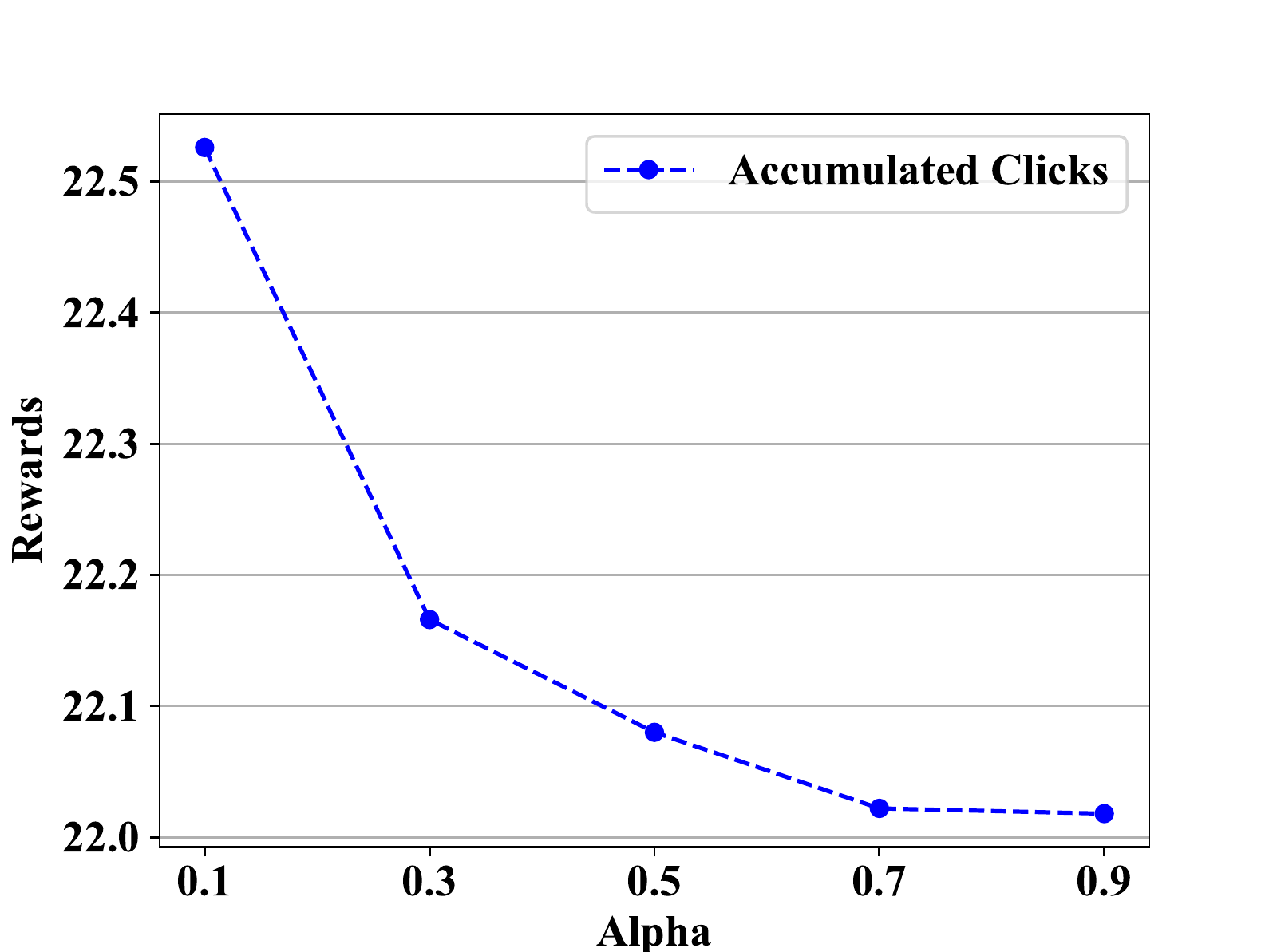}}
  \centerline{(b)}
\end{minipage}
\hfill
\begin{minipage}{0.32\linewidth}
  \centerline{\includegraphics[width=2.9cm]{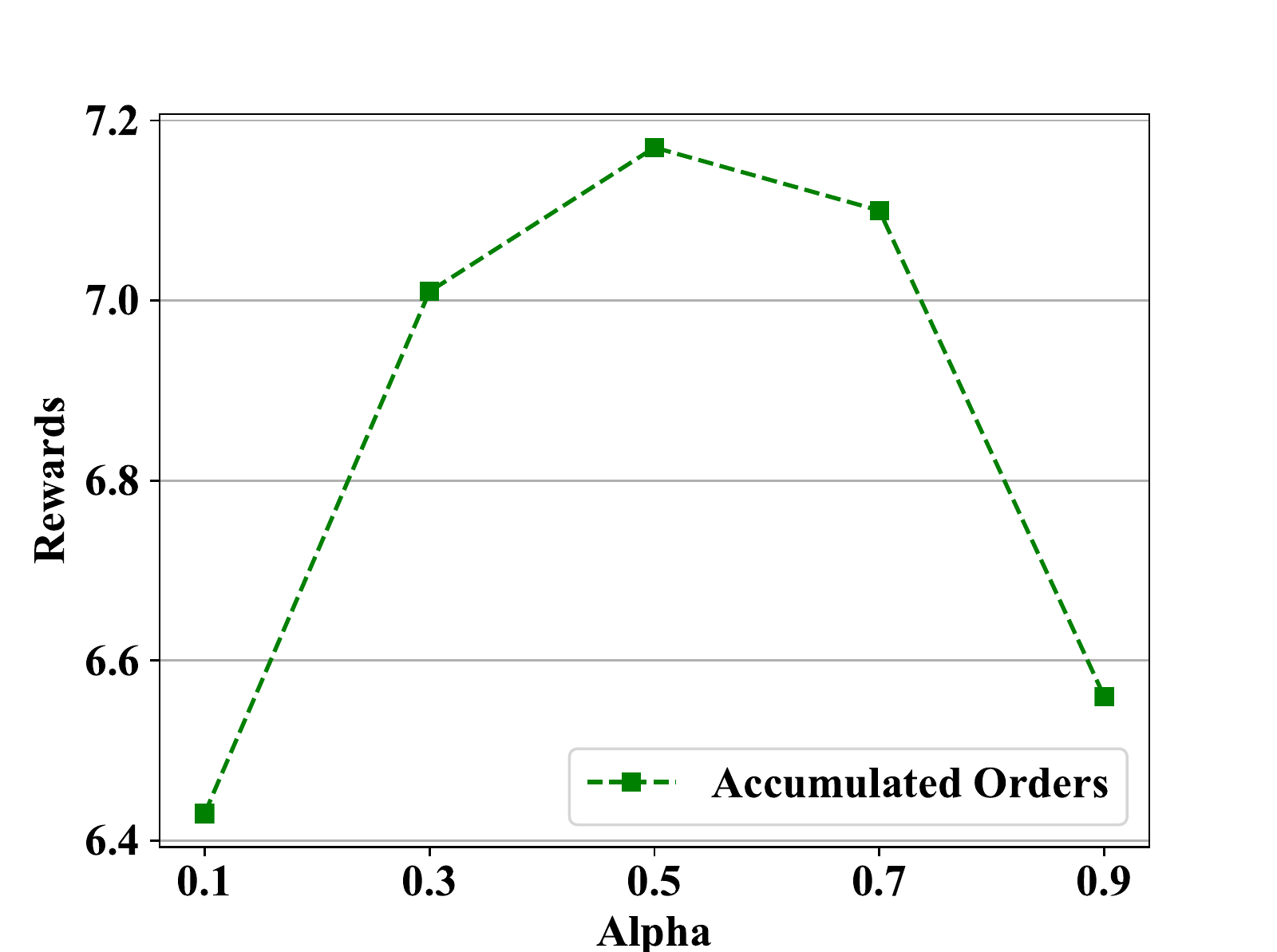}}
  \centerline{(c)}
\end{minipage}
\caption{Parameter sensitiveness of $\alpha$.}
\end{figure}

Figure 9 shows the parameter sensitivity of $\alpha$ in online recommendation task(long session). The performance for the recommendation achieves peak when $\alpha=0.5$. In other words, the high-level goals indeed improve the performance of the framework. It can be observed that as $\alpha$ increases, the cumulative clicks gradually decrease, indicating that the conversion information will have a negative impact on clicks due to their incomplete positive correlation. Choosing a suitable $\alpha$ can significantly improve the cumulative orders and total rewards. 

Figure 10 shows the parameter sensitivity of $M$ in online recommendation task(long session). The performance for the recommendation achieves peak when $M=2$. It can be observed that too many goals will cause the cumulative clicks to decrease and affect the overall performance of the framework. However, when the cumulative clicks have been greatly reduced when $M=3$ or 4, the cumulative orders are still higher than those of a single goal, which fully illustrates the promotion of multi-goals for the conversions.

\begin{figure}
\begin{minipage}{0.32\linewidth}
  \centerline{\includegraphics[width=3.0cm]{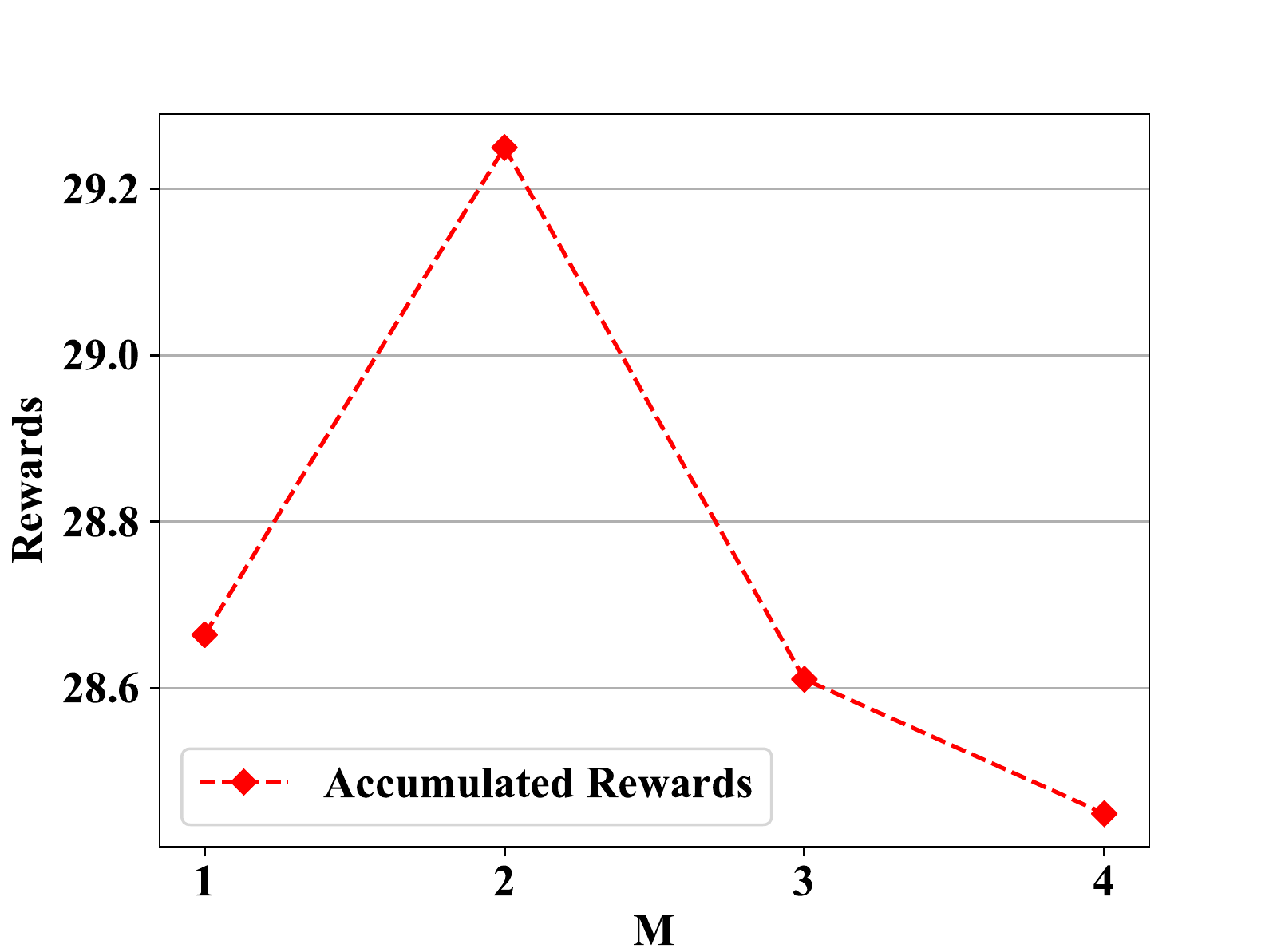}}
  \centerline{(a)}
\end{minipage}
\hfill
\begin{minipage}{0.32\linewidth}
  \centerline{\includegraphics[width=3.0cm]{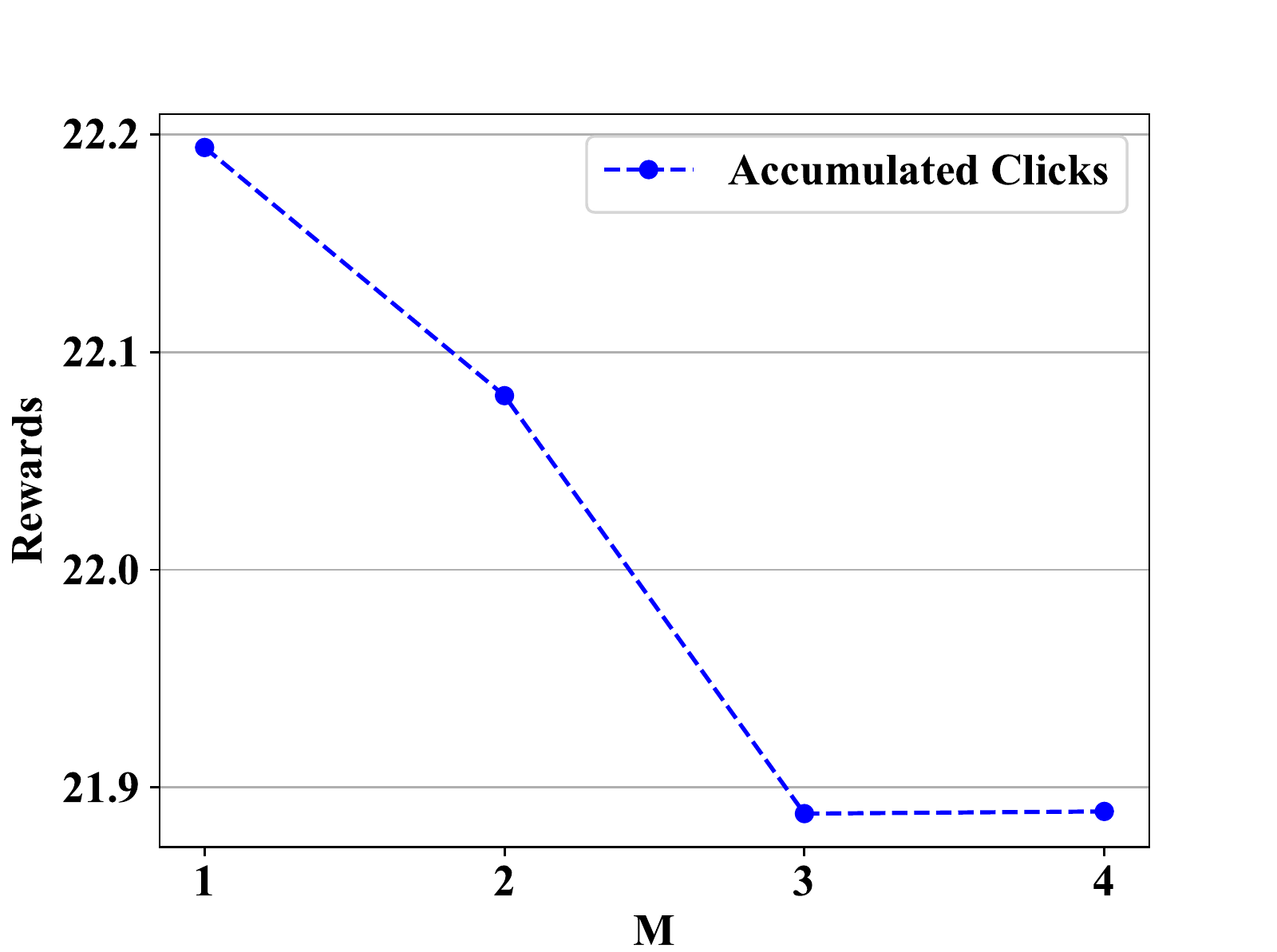}}
  \centerline{(b)}
\end{minipage}
\hfill
\begin{minipage}{0.32\linewidth}
  \centerline{\includegraphics[width=3.0cm]{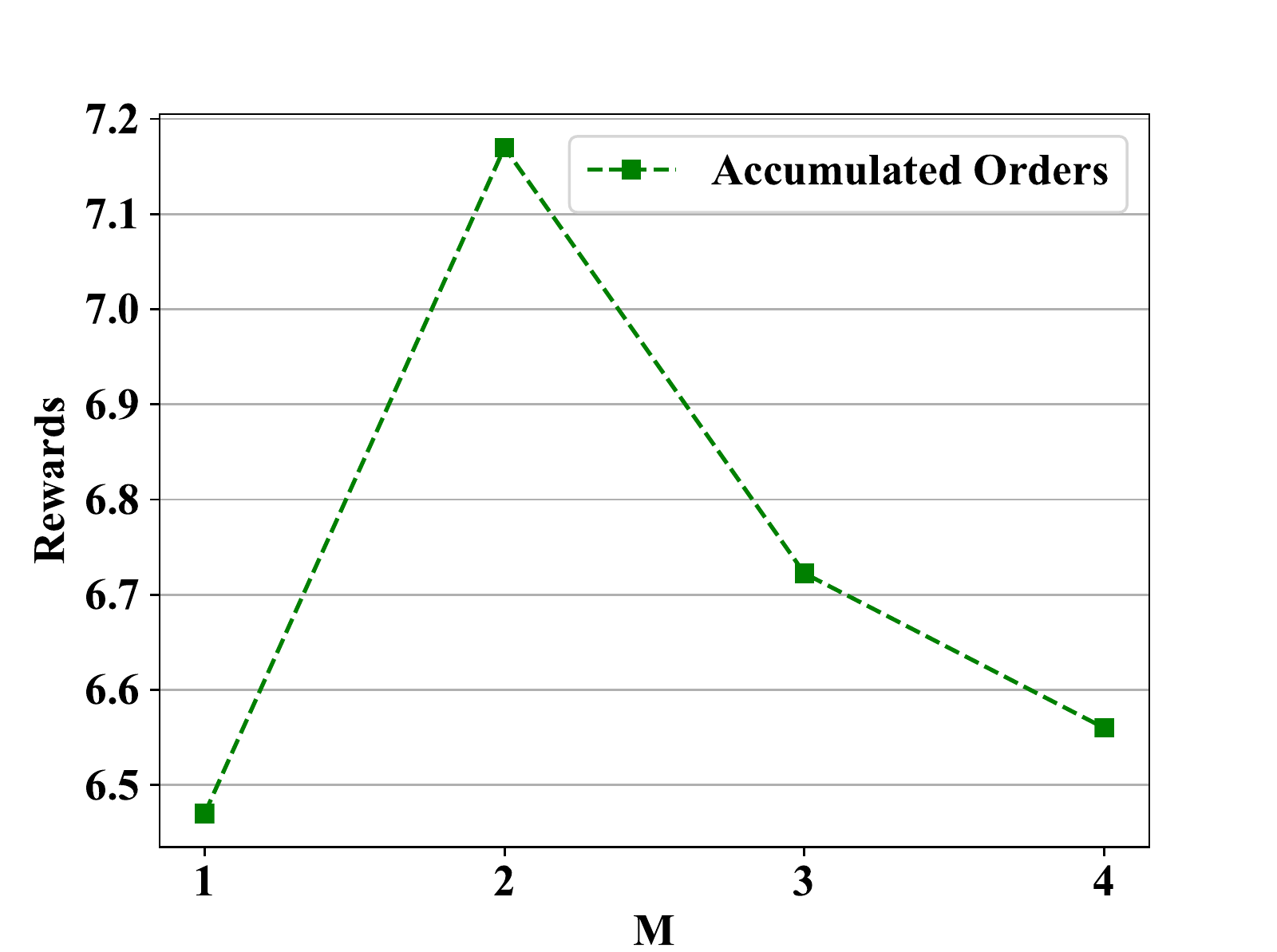}}
  \centerline{(c)}
\end{minipage}
\caption{Parameter sensitiveness of $M$.}
\end{figure}

\section{Related Work}
The recommendation algorithms can be roughly divided into three categories: traditional recommendation algorithms, deep learning based and reinforcement learning based recommendation algorithms. 
Firstly, traditional recommendation algorithms consists of collaborative filtering\cite{John1998}, content-based filtering\cite{Raymond2000},and hybrid methods\cite{Robin2002}.
Secondly, deep learning based recommendation algorithms have become the current mainstream recommendation methods. 
Deep learning methods can help to learn item embedding from sequences, image or graph information\cite{Paul2016}. It can also extract users' potential tastes\cite{Wu2016}, or improve the traditional methods directly\cite{Zhang2017}. 


Thirdly, reinforcement learning based recommendation algorithms are far more different from the above two categories. 
It models the recommending procedure as the interaction sequences between users(environment) and recommendation agent, and leverages reinforcement learning to automatically learn the optimal recommendation strategies. 
For instance, Li et al.\cite{Li2010} presented a contextual-bandit approach for personalized news article recommendation, in which a part of new items are exposed to balance exploration and exploitation.
Zhao et al.\cite{listwise,pagewise} proposed a novel page-wise recommendation framework based on reinforcement learning, which can optimize a page of items with proper display based on real-time feedback from users.

Deep hierarchical reinforcement learning is dedicated to expanding and combining existing reinforcement learning methods to solve more complex and difficult problems\cite{Richard1999,Andrew2003}. 
There is no doubt that the recommendation problem is such a problem. 
Recently, a goal-based hierarchical reinforcement learning framework\cite{Alexander2017,Ofir2018} has emerged, with high-level and low-level communicating through goals. 
However, as far as we know, there is no existing hierarchical reinforcement learning method for recommendation system.

\section{Conclusion}
In this paper, we propose a novel hierarchical reinforcement learning based recommendation framework, which consists of two components, i.e., high-level agent and low-level agent. The high-level agent tries to catch long-term sparse conversion signals, and automatically sets abstract multi-goals for the low-level agent, while the low-level agent follows different goals in different stage and interacts with real-time environment. The multiple high-level goals reduce the difficulty for the low-level agent to approach the high-level goals and accelerate the convergent rate of our proposed algorithm.
The experimental results based on a real-world e-commerce dataset demonstrate the effectiveness of the proposed framework. There are several interesting research directions.  Firstly, the low-level agent can be guided in other ways, such as a hidden state representing the long-term preference. Secondly, the framework is general, and more specific information can be used to improve the performance in specific tasks, such as category information of items, user profiles, etc. 



%
\bibliographystyle{ACM-Reference-Format}
\bibliography{sample-base}

%
\newpage
\appendix

\section{Discussions on Framework}

In practice, there are three basic difficulties in recommendation tasks: a) the number of users that recommendation platform have to serve is up to several hundred millions, and their preferences vary greatly; 
b) the number of items to be recommended increases rapidly and changes dynamically with time goes by, which means some items are deleted while some are added; 
c) it is time-consuming to select the optimal one from the set of alternative items.

In reinforcement learning, a) means a huge state space and b) means a huge and dynamic action space. 
In addition, The action value function is usually highly nonlinear, and many state-action pairs may not appear in the real trace such that it is hard to update their values.
Thus traditional reinforcement learning methods such as POMDP\cite{Guy2005} and Q-learning\cite{Nima2008} are not suitable because they cannot store massive data and handle complex relationships. 
Deep Q-Network(DQN)\cite{Volodymyr2013} is also not applicable, because the huge action space will greatly reduce the update speed of DQN.
Therefore, we must leverage deep reinforcement learning method\cite{Timothy2015}, using deep neural network as a nonlinear function approximation to approximate the policy and Q-value function simultaneously. Thus a fundamentally Actor-Critic architecture\cite{Richard1998} is needed.
In practice, it is not enough to represent items using only discrete indexes, because such representations have no semantic meaning and do not represent relationships between different items. 
A common practice is to extract the information of each item like sentences or images and embed them into a continuous abstract action space\cite{Omer2014}.

\section{Algorithm}

\subsection{Mapping Algorithm}
We proposed the mapping algorithm in Algorithm 1. The LActor generates a virtual-action $\hat{a}$(line 1), and selects the most similar item based on the cosine similarity(line 2). Finally, this item  is removed from the item embedding set(line 3), which prevents the same item is  recommended repeatedly in a session. Then the LActor recommends $a$ to user and receive immediate reward from user.

\begin{algorithm}[h]  
  \caption{Mapping Algorithm.}  
  \begin{algorithmic}[1]
    \Require User's low-level state $s^l$, item embedding set $I$.
    \Ensure Valid recommendation item $a$.
    \State Generate proto-action $\hat{a}$ according Eq.(13).
    \State Select the most similar item $a$ according Eq.(17).
    \State Remove item $a$ from $I$
    \State \Return $a$
  \end{algorithmic}  
\end{algorithm}  

\subsection{Online Training Algorithm}
\begin{algorithm}[htb]  
  \caption{Online Training Algorithm.}  
  \begin{algorithmic}[1]
    \State Initialize HActor $f_{\Theta^h_{\pi_i}}$, HCritic $Q_{\Theta^h_{\mu_i}}$, LActor $f_{\Theta^l_{\pi}}$, LCritic $Q_{\Theta^l_{\mu}}$ with random weights
    \State Initialize target network $f_{{\Theta^h_{\pi_i}}'}$, $Q_{{\Theta^h_{\mu_i}}'}$, $f_{{\Theta^l_{\pi}}'}$, $Q_{{\Theta^l_{\mu}}'}$ with weights $f_{{\Theta^h_{\pi_i}}'}\leftarrow f_{\Theta^h_{\pi_i}}, Q_{{\Theta^h_{\mu_i}}'} \leftarrow Q_{\Theta^h_{\mu_i}}, f_{{\Theta^l_{\pi}}'} \leftarrow f_{\Theta^l_{\pi}}, Q_{{\Theta^l_{\mu}}'} \leftarrow Q_{\Theta^l_{\mu}}$
    \State Initialize the capacity of high-level and low-level replay buffer $D^h, D^l$
    \For{$session \in [1, G]$}
      \State Initialize clock $t\leftarrow 0$
      \State Receive initial high-level and low-level state $s^h_t, s^l_t$
      \While{$t < T$}  
        \State Stage 1. Transition Generating Stage
        \If{$t\equiv 0(mod~c)$}
          \State Generate a set of goals $g^{1:M}_t$ according to Eq.(5)
        \Else
          \State $g^{1:M}_t \leftarrow g^{1:M}_{t-1}$
        \EndIf
        \State Select an action $a_t$ according to Alg.1
        \State Execute action $a_t$ and observe external reward $r^{ex}_t$ 
        \State New high-level and low-level state $s^h_{t+1}, s^l_{t+1}$
        \State Store low-level transition $(s^l_t,g^{1:M}_t,a_t,r^{ex}_t,s^l_{t+1})$ in $D^l$
        \State $t\leftarrow t+1$
        \If{$t\equiv 0(mod~c)$} 
          \State Collect the recent $c$ external rewards $r^{ex}_{t-c:t-1}$ and store high-level transition $(s^h_{t-c},g^{1:M}_{t-c},r^{ex}_{t-c:t-1},s^h_{t})$ in $D^h$
        \EndIf
        \State Stage 2. Parameter updating stage
        \State Sample mini-batch of $N^h$ high-level transitions 
        $(s^h_{t-c},g^{1:M}_{t-c},r^{ex}_{t-c:t-1},s^h_{t})$ from $D^h$
        \State Update HCritic, HActor according to Eq.(17)(18)
        \State Sample mini-batch of $N^l$ low-level transitions $(s^l_t,g^{1:M}_t,a_t,r^{ex}_t,s^l_{t+1})$ from $D^l$
        \State Update LCritic, LActor according to Eq.(20)(21)
        \State Update the target networks:
        $$f_{{\Theta^h_{\pi_i}}'}\leftarrow \tau f_{\Theta^h_{\pi_i}} + (1-\tau)f_{{\Theta^h_{\pi_i}}'}$$
        $$Q_{{\Theta^h_{\mu_i}}'}\leftarrow \tau Q_{\Theta^h_{\mu_i}} + (1-\tau)Q_{{\Theta^h_{\mu_i}}'}$$
        $$f_{{\Theta^l_{\pi}}'}\leftarrow \tau f_{\Theta^l_{\pi}} + (1-\tau)f_{{\Theta^l_{\pi}}'}$$
        $$Q_{{\Theta^l_{\mu}}'}\leftarrow \tau Q_{\Theta^l_{\mu}} + (1-\tau)Q_{{\Theta^l_{\mu}}'}$$
      \EndWhile
    \EndFor 
  \end{algorithmic}  
\end{algorithm} 
The online training algorithm for the proposed recommendation framework based on hierarchical reinforcement learning is presented in Algorithm 2. In each iteration, there are two stages: 1) transition generation stage(lines 8-21); 2) parameter updating stage(lines 22-27). For transition generating stage: given the current high-level state $s^h_t$ and low-level state $s^l_t$, the HRA first generates a set of goals $g_t^{1:M}$ when $t\equiv 0(mod~c)$ and conveys them to the LRA(line 10); the LRA recommends an item $a_t$ according to Algorithm 1(line 14); next, the RA observes the external reward $r^{ex}_t$ (line 15)and updates the high-level and low-level state to ${s^h}', {s^l}'$(line 16); then, the LRA stores transitions $(s^l_t,g^{1:M}_t,a_t,r^{ex}_t,s^l_{t+1})$ in the low-level replay buffer $D^l$(line 17); and finally, after $c$ time steps(when $t\equiv 0 (mod~c)$ again), the LRA collects the recent $c$ external rewards $r^{ex}_{t-c:t-1}$ and conveys them to HRA, the HRA will store the transition $(s^h_{t-c},g^{1:M}_{t-c},r^{ex}_{t-c:t-1},s^h_{t})$ in the high-level replay buffer $D^h$(line 20). For parameter updating stage: the HRA samples mini-batch of transitions $(s^h_{t-c},g^{1:M}_{t-c},r^{ex}_{t-c:t-1},s^h_{t})$ from $D^h$ and updates parameters of HActor and HCritic, while the LRA samples mini-batch of transitions $(s^l_t,g^{1:M}_t,a_t,r^{ex}_t,s^l_{t+1})$ from $D^l$ and updates parameters of LActor and LCritic(lines 23-27), following a standard DDPG procedure\cite{Timothy2015}.

In the algorithm, we introduce widely used techniques to train our framework. For instance, we use a technique known as experience replay\cite{Lin1993}(lines 23,25), and introduce separated evaluation and target networks\cite{Volodymyr2013}(lines 2,27), which can help smooth the learning and avoid the divergence of parameters. For the soft updates of target networks(lines 27), we set $\tau=0.01$.
\section{The Test Procedure}
After the training procedure, the proposed recommendation framework learns parameters $f_{\Theta^h_{\pi_i}}$, $Q_{\Theta^h_{\mu_i}}$, $f_{\Theta^l_{\pi}}$, $Q_{\Theta^l_{\mu}}$. Here we formally present the test procedure of the proposed framework. We design two methods: 1) Online test: to test the framework in online environment where the agents interact with users and receive real-time feedback for the recommended items from users; 2) Offline test: to test the framework based on user's historical logs.
\subsection{Online Test}
The online test algorithm in one recommendation is presented in Algorithm 3. The online test procedure is similar with the transition generating stage in Algorithm 2. In each iteration of the recommendation session, given the current low-level state $s^l_t$, the LRA recommends an item $a_t$ to user following policy $f_{\Theta^l_{\pi}}$(line 4). Then the LRA observes the external reward $r^{ex}_t$ from user(line 5) and updates the low-level state to $s^l_{t+1}$(line 6).
\begin{algorithm}[htb]  
  \caption{Online Test Algorithm.}  
  \begin{algorithmic}[1]
    \State Initialize LActor the trained parameters $\Theta^l_{\pi}$
    \State Receive initial low-level state $s^l_0$
    \For{$t \in [0, T]$}  
      \State Select an action $a_t$ according to Alg.1
      \State Execute action $a_t$ and observe external reward $r^{ex}_t$ 
      \State New low-level state $s^l_{t+1}$
    \EndFor 
  \end{algorithmic}  
\end{algorithm} 
\subsection{Offline Test}
The intuition of the offline test method is that, for a given recommendation(offline data), the LRA reranks the items in this session. If the proposed framework works well, the clicked/ordered items in this session will be ranked at the top of the new list. The reason why LRA only reranks items in this session rather than items in the while item space is that for the offline dataset, we only have the ground truth rewards of the existing items in this session. The offine test algorithm in one recommendation session is presented in Algorithm 4. In each iteration of an offline test recommendation session, given the low-level state $s^l_t$(line 2), the LRA recommends an item $a_t$ following policy $f_{\Theta^l_{\pi}}$(line 4). And then, we add $a_t$ into new recommendation list $L$(line 5), and record $a_t$'s external reward $r^{ex}_t$ from user's historical data(line 6). Then we update the low-level state to $s^l_{t+1}$(line 7). Finally, we remove $a_t$ from the item set $I$ of the current session(line 8).
\begin{algorithm}[htb]  
  \caption{Offline Test Algorithm.}  
  \begin{algorithmic}[1]
    \Require Item embedding set $I=\{e_1, e_2, \cdots, e_N\}$ and corresponding external reward set $R^{EX}=\{r^{ex}_1, r^{ex}_2, \cdots, r^{ex}_N\}$.
    \Ensure Recommendation list $L$ with new order.
    \State Initialize LActor the trained parameters $\Theta^l_{\pi}$
    \State Receive initial low-level state $s^l_0$
    \While{$|I|>0$}  
      \State Select an action $a_t$ according to Alg.1
      \State Add action $a_t$ into the end of $L$
      \State Record external reward $r^{ex}_t$ from user's historical data
      \State New low-level state $s^l_{t+1}$
      \State Remove $a_t$ from $I$
    \EndWhile 
  \end{algorithmic}  
\end{algorithm}

\section{Statistics on the dataset}
Long tail data is filtered in this dataset:
\begin{table}[h]
  \caption{Statistics on the dataset(Year:2018)}
  \label{tab:freq}
  \begin{tabular}{cccccc}
    \toprule
    Dataset& Date& Samples& SKU& Clicks& Orders\\
    \midrule
    Train\_set & Aug.11th& 8,596,852& 553,156& 843,249& 46,022\\
    Test\_set & Aug.12th& 2,231,651& 287,689& 218,053& 10,552\\
  \bottomrule
\end{tabular}
\end{table}

\section{Parameter Sensitivity in Short Session}
The parameter sensitivity of $\alpha$ and $M$ in online recommendation task(short session) are shown in Figure 11,12.
\begin{figure}[h]
\begin{minipage}{0.32\linewidth}
  \centerline{\includegraphics[width=3.0cm]{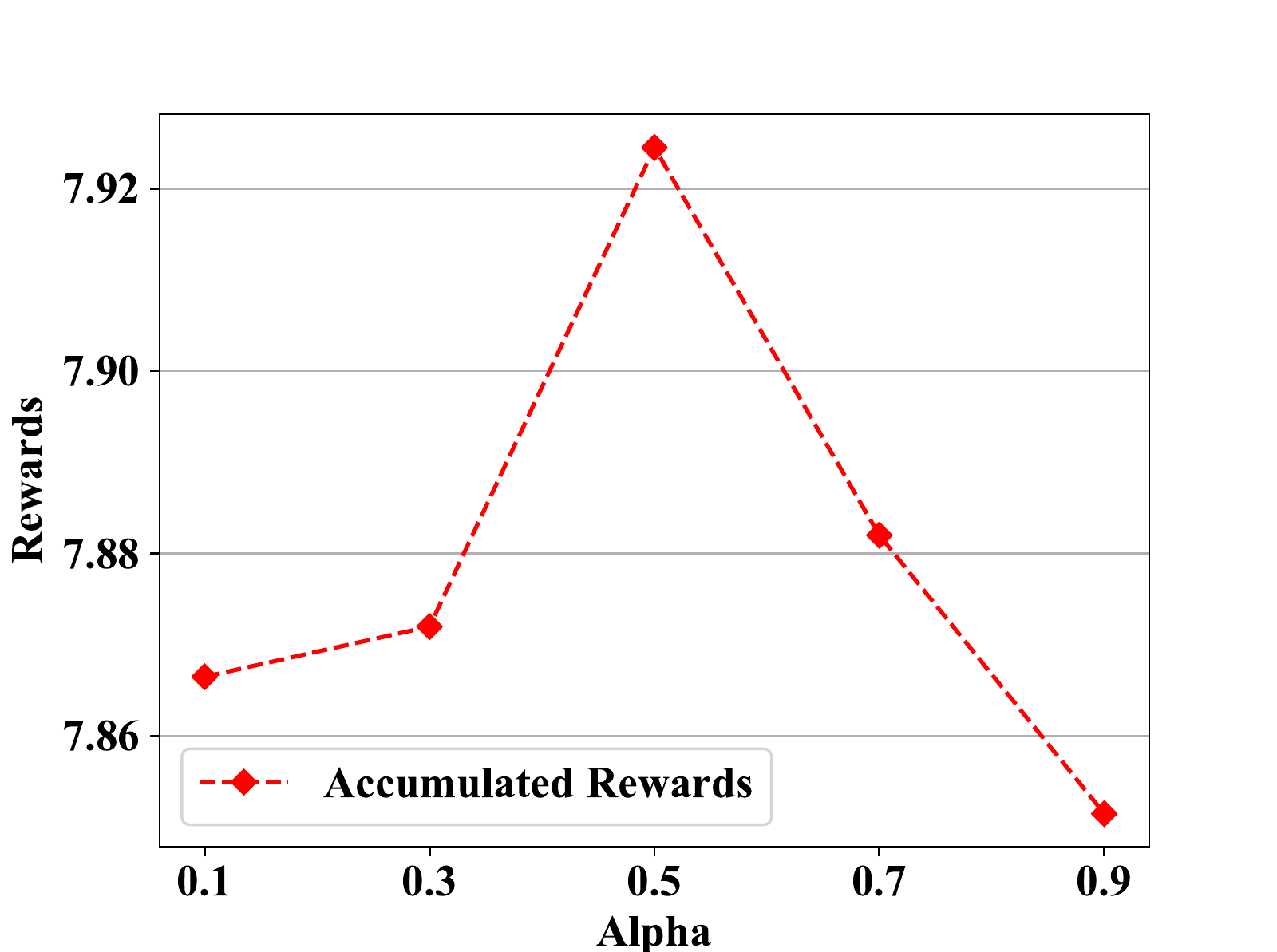}}
  \centerline{(a)}
\end{minipage}
\hfill
\begin{minipage}{0.32\linewidth}
  \centerline{\includegraphics[width=3.0cm]{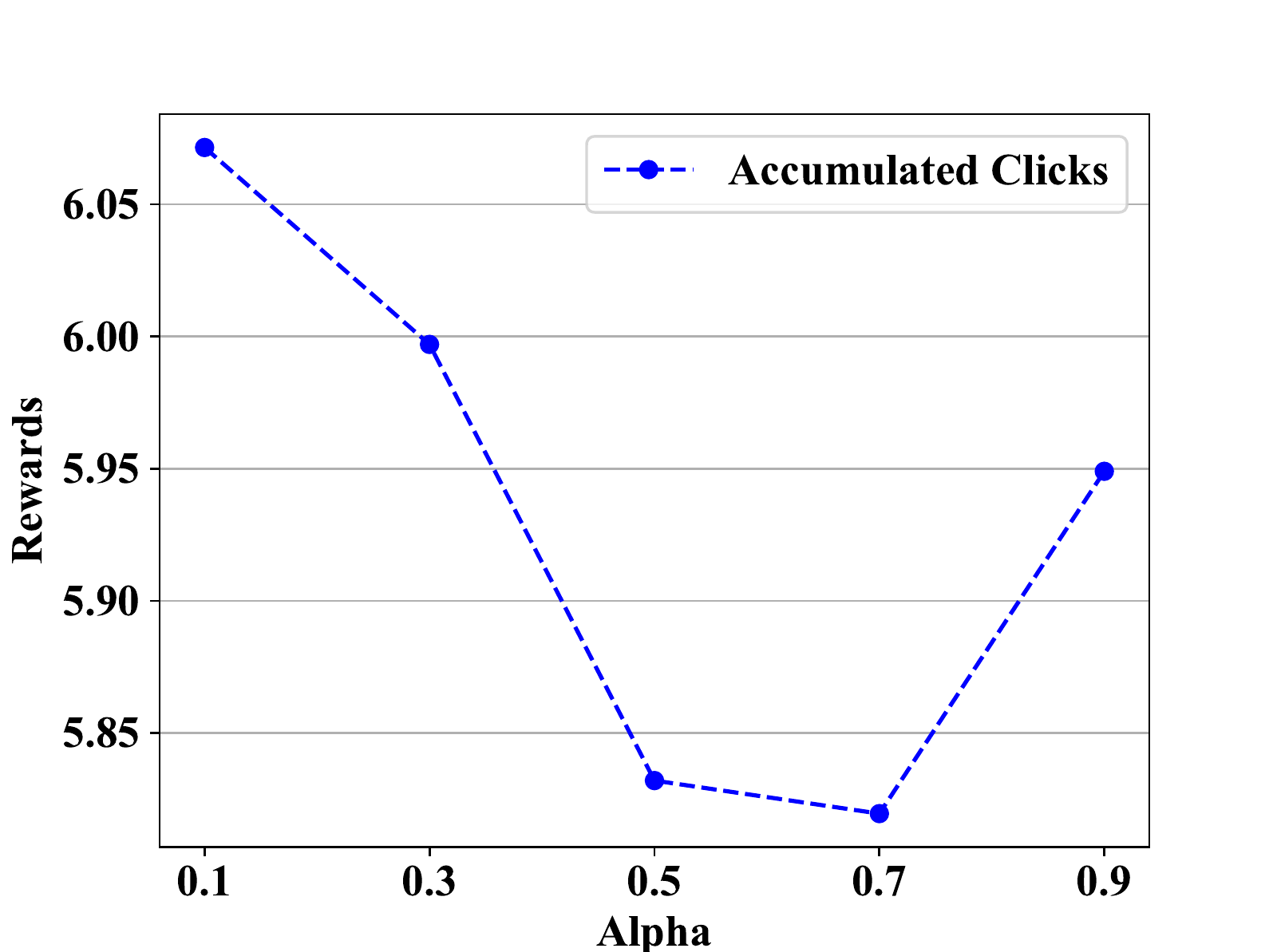}}
  \centerline{(b)}
\end{minipage}
\hfill
\begin{minipage}{0.32\linewidth}
  \centerline{\includegraphics[width=3.0cm]{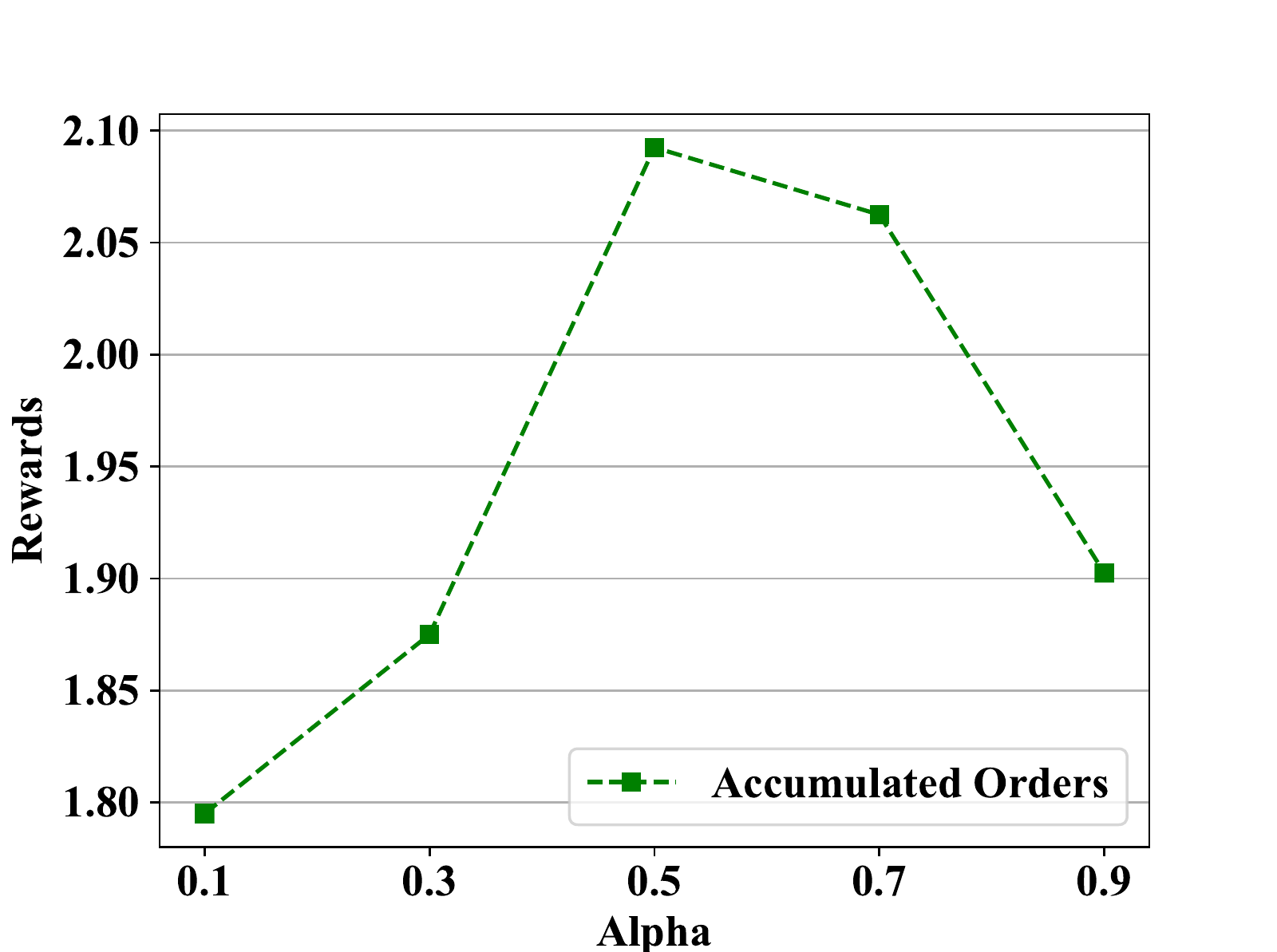}}
  \centerline{(c)}
\end{minipage}
\caption{Parameter sensitiveness of $\alpha$ in short session.}
\end{figure}

\begin{figure}[h]
\begin{minipage}{0.32\linewidth}
  \centerline{\includegraphics[width=3.0cm]{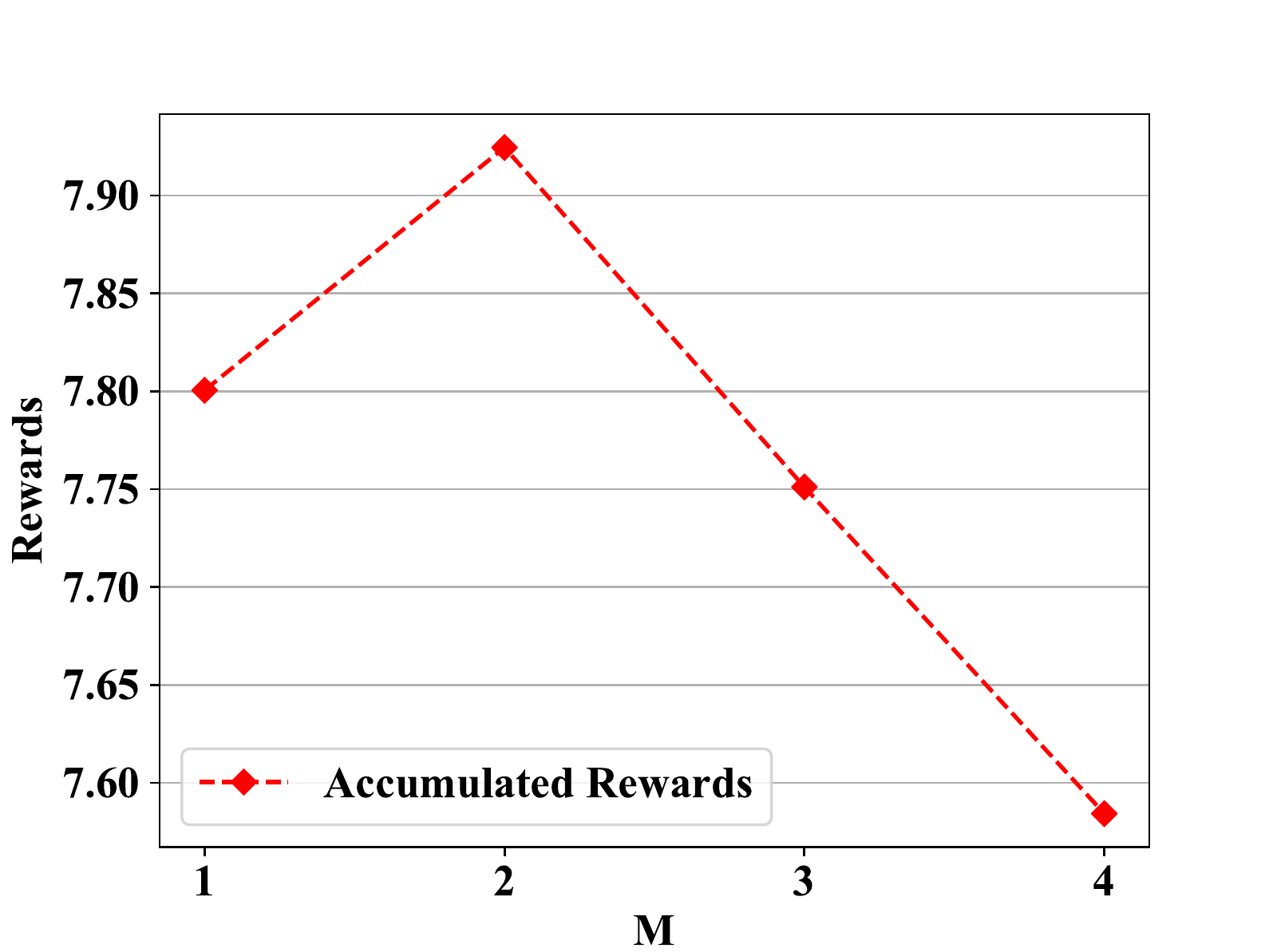}}
  \centerline{(a)}
\end{minipage}
\hfill
\begin{minipage}{0.32\linewidth}
  \centerline{\includegraphics[width=3.0cm]{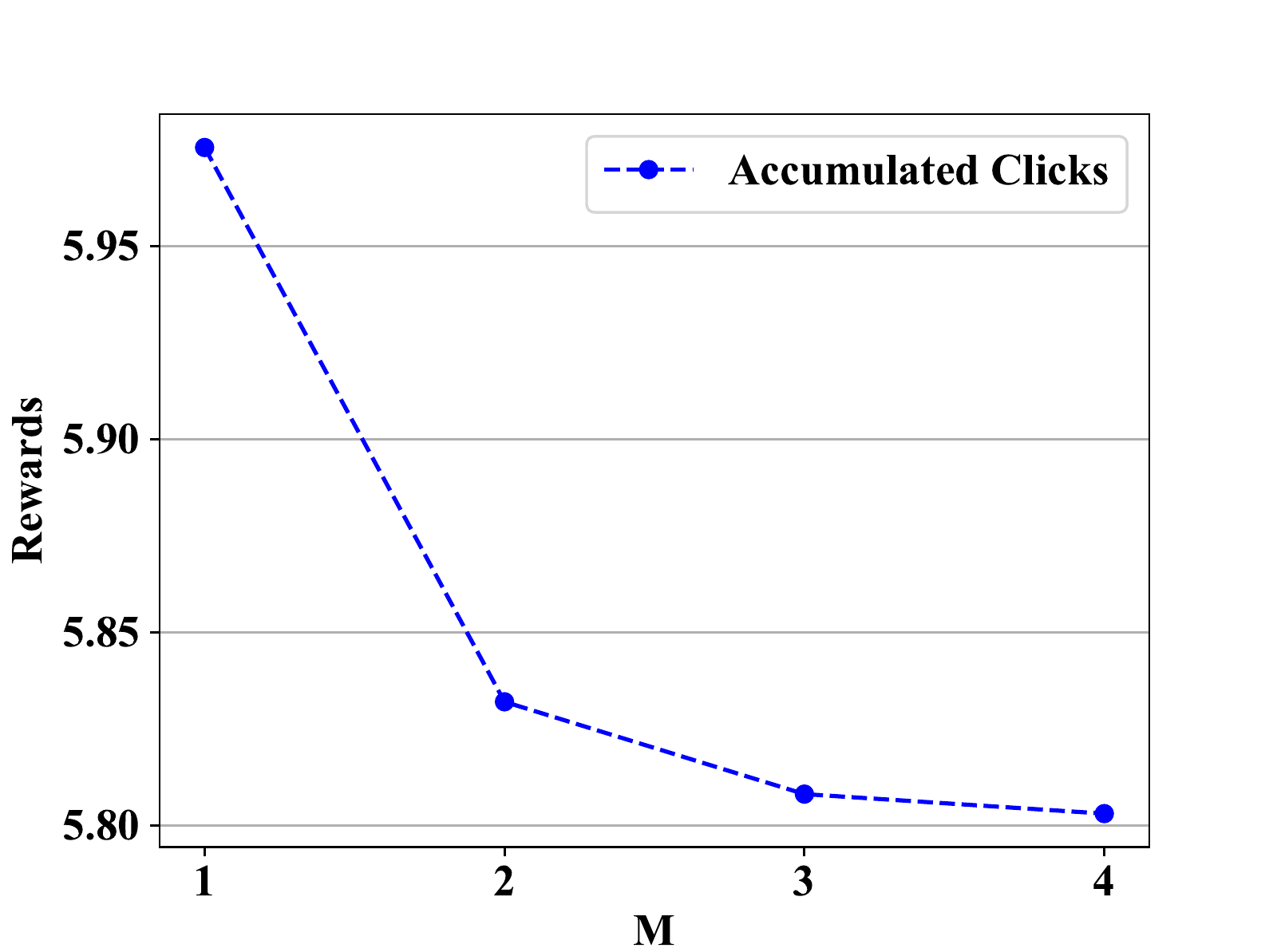}}
  \centerline{(b)}
\end{minipage}
\hfill
\begin{minipage}{0.32\linewidth}
  \centerline{\includegraphics[width=3.0cm]{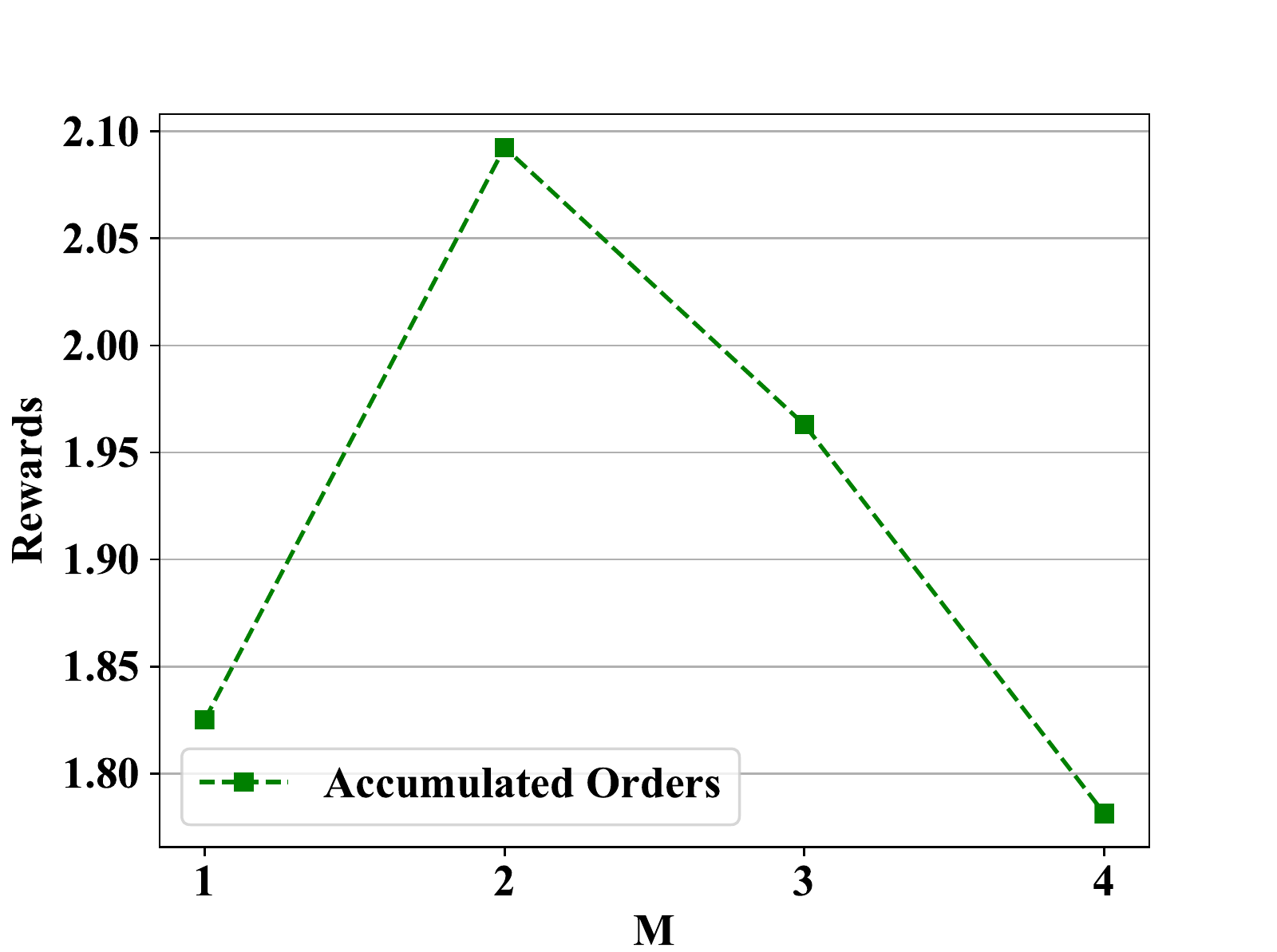}}
  \centerline{(c)}
\end{minipage}
\caption{Parameter sensitiveness of $M$ in short session.}
\end{figure}
\end{document}